\RequirePackage{snapshot}

\documentclass{article}

\usepackage[final]{neurips_2018}
\usepackage{mycustom}

\usepackage[utf8]{inputenc} 
\usepackage[T1]{fontenc}    
\usepackage{hyperref}       
\usepackage{url}            
\usepackage{booktabs}       
\usepackage{amsfonts}       
\usepackage{nicefrac}       
\usepackage{microtype}      

\title{\Gilbo/: One Metric to Measure Them All}

\author{Alexander A.~Alemi\footnotemark[1], Ian Fischer\thanks{Authors contributed equally.}\\
Google AI \\
\texttt{\{alemi,iansf\}@google.com}}

\begin{document}

\maketitle

	\begin{abstract}
	We propose a simple, tractable lower bound on the mutual information
	contained in the joint generative density of any latent variable
	generative model: the \gilbo/ (\textit{Generative Information Lower
	BOund}).  It offers a data-independent measure of the complexity of the
	learned latent variable description, giving the log of the effective
	description length.  It is well-defined for both \vae/s and \gan/s.  We
	compute the \gilbo/ for 800 \gan/s and \vae/s each trained on four
	datasets (\textsc{mnist}, Fashion\textsc{mnist}, \textsc{cifar}-10 and
	CelebA) and discuss the results.
\end{abstract}

}%

	\section{Introduction}

\Gan/s~\citep{gan} and \vae/s~\citep{vae} are the most popular latent variable
generative models because of their relative ease of training and high
expressivity.  However \emph{quantitative} comparisons across different
algorithms and architectures remains a challenge.
\Vae/s are generally measured using the \elbo/, which measures their fit to
data.  Many metrics have been proposed for \gan/s, including the \inception/
score~\citep{inceptionscore}, the \fid/ score~\citep{fid}, independent
Wasserstein critics~\citep{iwc}, birthday paradox testing~\citep{ganbirthday},
and using Annealed Importance Sampling to evaluate log-likelihoods~\citep{wu2016quantitative},
among others.

Instead of focusing on metrics tied to the data distribution, we believe a
useful additional independent metric worth consideration is the complexity of
the trained generative model. Such a metric would help answer questions related
to overfitting and memorization, and may also correlate well with sample quality.
To work with both \gan/s and \vae/s our metric should not require a tractable joint
density $p(x,z)$.
To address these desiderata, we propose the \gilbo/.
}%

	\section{\Gilbo/: Generative Information Lower BOund}

A symmetric, non-negative, reparameterization independent measure of the
information shared between two random variables is given by the mutual
information:
\begin{equation}
	I(X;Z) = \iint dx\, dz\, p(x,z) \log \frac{p(x,z)}{p(x)p(z)} = \int dz\, p(z) \int dx\, p(x|z) \log \frac{p(z|x)}{p(z)} \geq 0 .
	\label{eqn:mutinfo}
\end{equation}
$I(X;Z)$ measures how much information (in nats) is learned about
one variable given the other.
As such it is a measure of the complexity of the generative model.
It can be interpreted (when converted to bits) as the reduction in the number of yes-no questions needed to
guess $X=x$ if you observe $Z=z$ and know $p(x)$, or vice-versa.
It gives the log of the \emph{effective description length} of the generative model.
This is roughly the log of the number of distinct sample pairs~\citep{tishby}.
$I(X;Z)$ is well-defined even for continuous distributions.
This contrasts with the continuous entropy $H(X)$ of the marginal distribution,
which is not reparameterization independent~\citep{content}.
$I(X;Z)$ is intractable due to the presence of $p(x) = \int dz\, p(z) p(x|z)$, but we can derive a tractable variational lower bound~\citep{agakov}:
\begin{align}
  I(X;Z) &= \iint dx\, dz\, p(x,z) \log \frac{p(x,z)}{p(x)p(z)} \\
  &= \iint dx\, dz\, p(x,z) \log \frac{p(z|x)}{p(z)} \\
  &\ge \iint dx\, dz\, p(x,z) \log p(z|x) - \int dz\, p(z) \log p(z) - \operatorname{KL}[p(z|x)||e(z|x)] \\
  &= \int dz\, p(z) \int dx\, p(x|z) \log \frac{e(z|x)}{p(z)} = \mathbb{E}_{p(x,z)} \left[ \log \frac{e(z|x)}{p(z)} \right] \equiv \textrm{\gilbo/} \leq I(X;Z)
	\label{eqn:gilbo}
\end{align}
We call this bound the \gilbo/ for \textit{Generative Information Lower BOund}.
It requires learning a tractable variational approximation to the intractable posterior
$p(z|x) = p(x,z)/p(x)$, termed $e(z|x)$ since it acts as an \emph{encoder}
mapping from data to a prediction of its associated latent variables.\footnote{%
  Note that a \emph{new} $e(z|x)$ is trained for both \gan/s and \vae/s.
  \Vae/s do not use their own $e(z|x)$, which would also give a valid lower bound.
  In this work, we train a new $e(z|x)$ for both to treat both model classes uniformly.
  We don't know if using a new $e(z|x)$ or the original would tend to result in a tighter bound.
}
As a variational approximation, $e(z|x)$ depends on some parameters, $\theta$, which we elide in the notation.

The encoder $e(z|x)$ performs a regression for the inverse of the \gan/ or
\vae/ generative model, approximating the latents that gave rise to an observed
sample.  This encoder should be a tractable distribution, and must respect the
domain of the latent variables, but does not need to be reparameterizable as no
sampling from $e(z|x)$ is needed during training.  We suggest the use of
$(-1,1)$ remapped Beta distributions in the case of uniform latents, and
Gaussians in the case of Gaussian latents.  In either case, training the
variational encoder consists of simply generating pairs of $(x,z)$ from the
trained generative model and maximizing the likelihood of the encoder to
generate the observed $z$, conditioned on its paired $x$, divided by the
likelihood of the observed $z$ under the generative model's prior, $p(z)$.  For the GANs
in this study, the prior was a fixed uniform distribution, so the $\log p(z)$
term contributes a constant offset to the variational encoder's likelihood.
Optimizing the \gilbo/ for the parameters of the encoder gives a lower bound on
the true generative mutual information in the \gan/ or \vae/. Any failure to
converge or for the approximate encoder to match the true distribution does not
invalidate the bound, it simply makes the bound looser.

The \gilbo/ contrasts with the \emph{representational mutual
information} of \vae/s defined by the data and encoder, which motivates \vae/
objectives~\citep{brokenelbo}. For \vae/s, both lower and
upper variational bounds can be defined on the representational joint
distribution ($p(x)e(z|x)$).
These have demonstrated their utility for cross-model comparisons.
However, they require a tractable posterior, preventing their use with
most \gan/s.
The \gilbo/ provides a theoretically-justified and dataset-independent
metric that allows direct comparison of \vae/s and \gan/s.

The \gilbo/ is entirely independent of the \emph{true} data, being
purely a function of the generative joint distribution.
This makes it distinct from other proposed metrics like estimated marginal log
likelihoods (often reported for \vae/s and very expensive to estimate for \gan/s)~\citep{wu2016quantitative}\footnote{%
  Note that \citet{wu2016quantitative} is complementary to our work, providing both upper and lower bounds on the log-likelihood.
  It is our opinion that their estimates should also become standard practice when measuring \gan/s and \vae/s.
},
an independent Wasserstein critic~\citep{iwc},
or the common \inception/~\citep{inceptionscore} and \fid/~\citep{fid} scores which attempt to measure how
well the generated samples match the observed true data samples. Being independent of data,
the \gilbo/ does not directly measure sample quality, but extreme values (either low or high)
correlate with poor sample quality, as demonstrated in the experiments below.

Similarly, in~\citet{im2018quantitatively}, the authors propose using various \gan/ training objectives to quantitatively measure the performance of \gan/s on their own generated data.
Interestingly, they find that evaluating \gan/s on the same metric they were trained on gives paradoxically weaker performance
-- an \textsc{ls-gan} appears to perform worse than a Wasserstein \gan/ when evaluated with the least-squares metric, for example, even though the \textsc{ls-gan} otherwise outperforms the \textsc{wgan}.
If this result holds in general, it would indicate that using the \gilbo/ during training might result in less-interpretable evaluation \gilbo/s.
We do not investigate this hypothesis here.

Although the \gilbo/ doesn't directly reference the dataset,
the dataset provides useful signposts.
First is at $\log C$, the number of distinguishable classes in the data.
If the \gilbo/ is lower than that, the model has almost certainly failed to
learn a reasonable model of the data.
Another is at $\log N$, the number of training points. A \gilbo/ near this value may indicate that the model
has largely memorized the training set, or that the model's capacity happens
to be constrained near the size of the training set.
At the other end is the entropy of the data itself ($H(X)$)
taken either from a rough estimate, or from the best achieved data log likelihood
of any known generative model on the data.
Any reasonable generative model should have a \gilbo/ no higher than this value.

Unlike other metrics, \gilbo/ does not monotonically map to quality of the generated output.  Both
extremes indicate failures.  A vanishing \gilbo/ denotes a generative model with
vanishing complexity, either due to independence of the latents and samples, or
a collapse to a small number of possible outputs. A diverging \gilbo/
suggests over-sensitivity to the latent variables.

In this work, we focus on variational approximations to the generative information.
However, other means of estimating the \gilbo/ are also valid.
In~\Cref{sec:precision} we explore a computationally-expensive method to find a very tight bound.
Other possibilities exist as well, including the recently proposed \textit{Mutual Information Neural Estimation}~\citep{belghazi2018mutual} and \textit{Contrastive Predictive Coding}~\citep{cpc}.
We do not explore these possibilities here, but any valid estimator of the mutual information can be used for the same purpose.

}%

	\section{Experiments}

We computed the \gilbo/ for each of the 700 \gan/s and 100 \vae/s tested in
\citet{gansequal} on the \textsc{mnist}, Fashion\textsc{mnist}, \textsc{cifar} and CelebA datasets in their \emph{wide range}
hyperparameter search.  This allowed us to compare \fid/ scores and \gilbo/
scores for a large set of different \gan/ objectives on the same architecture.
For our encoder network, we duplicated the discriminator, but adjusted the
final output to be a linear layer predicting the $64 \times 2 = 128$ parameters
defining a $(-1, 1)$ remapped Beta distribution (or Gaussian in the case of the
\vae/) over the latent space. We used a Beta since all of the \gan/s were
trained with a $(-1, 1)$ 64-dimensional uniform distribution.  The parameters
of the encoder were optimized for up to 500k steps with
\textsc{adam}~\citep{adam} using a scheduled multiplicative learning rate
decay. We used the same batch size (64) as in the original training. Training time
for estimating \gilbo/ is comparable to doing \fid/ evaluations (a few minutes) on the small datasets (\textsc{mnist}, Fashion\textsc{mnist}, \textsc{cifar}), or over 10 minutes for larger datasets and models (CelebA).

\begin{figure}[!t]
	\centering
	\subfloat[\fid/]{\includegraphics[width=0.31\textwidth]{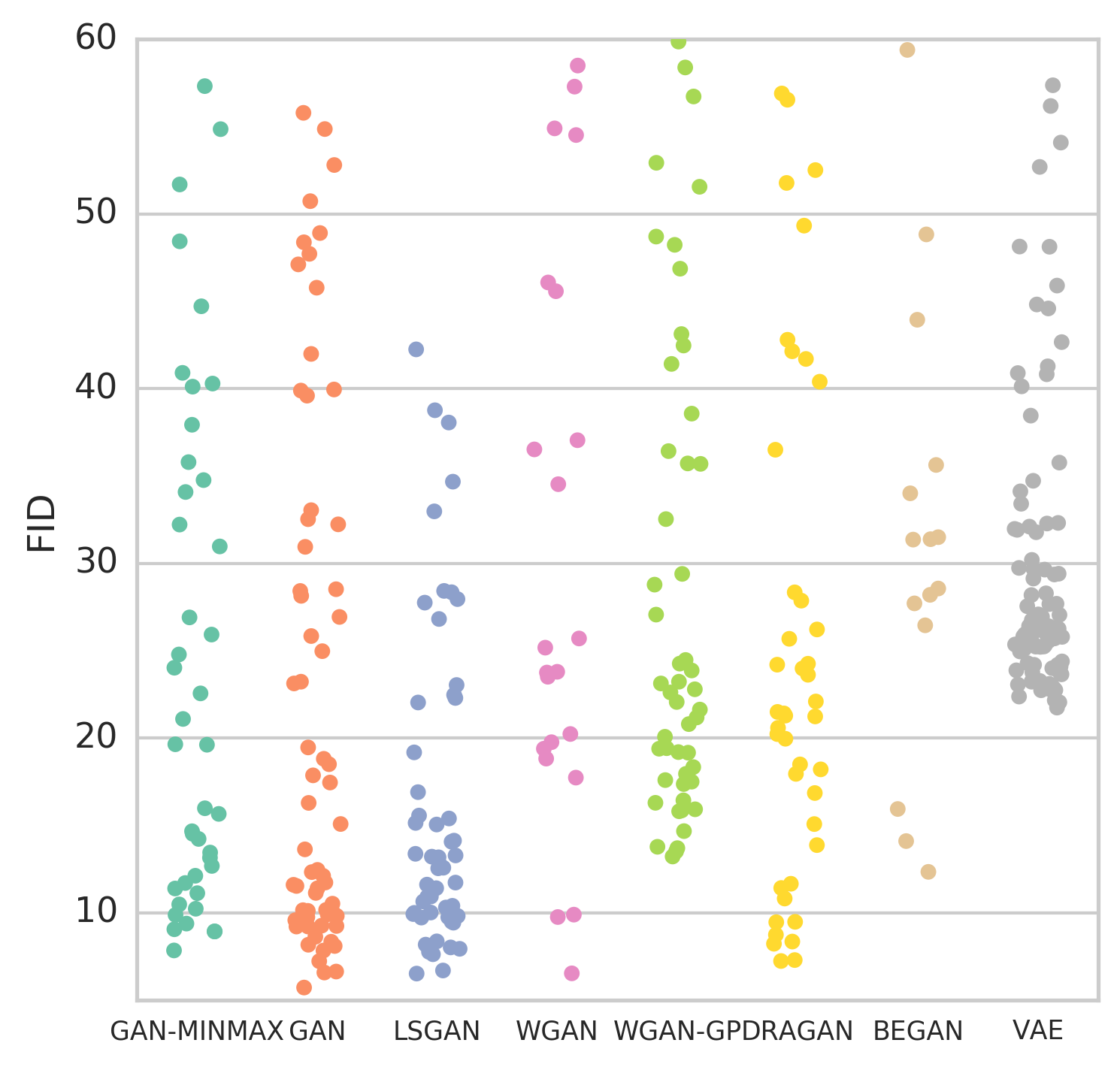}\label{fig:fidbox}}
	\subfloat[\gilbo/]{\includegraphics[width=0.31\textwidth]{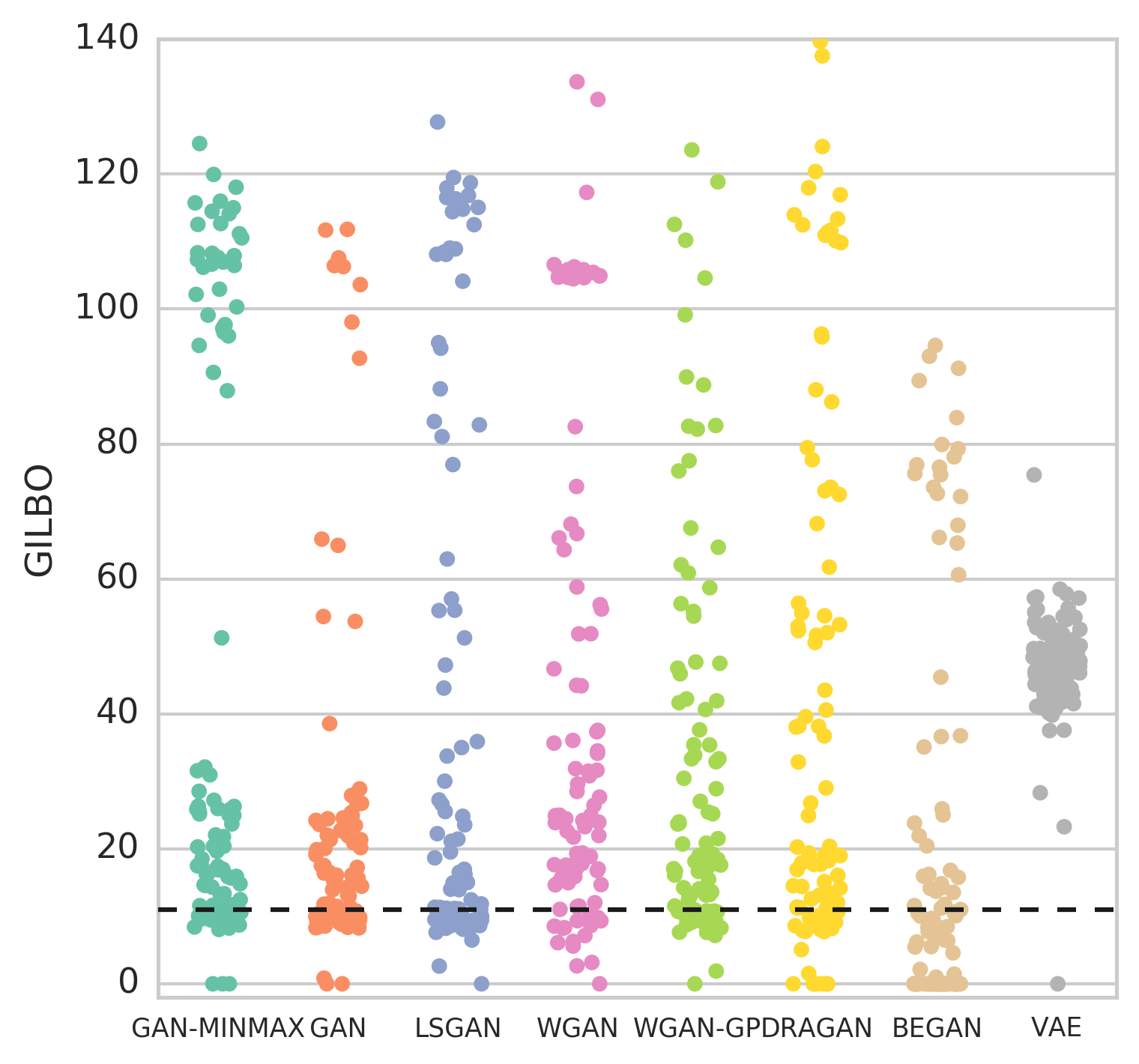}\label{fig:gilbobox}}
	\subfloat[\gilbo/ vs \fid/]{\includegraphics[width=0.3149\textwidth]{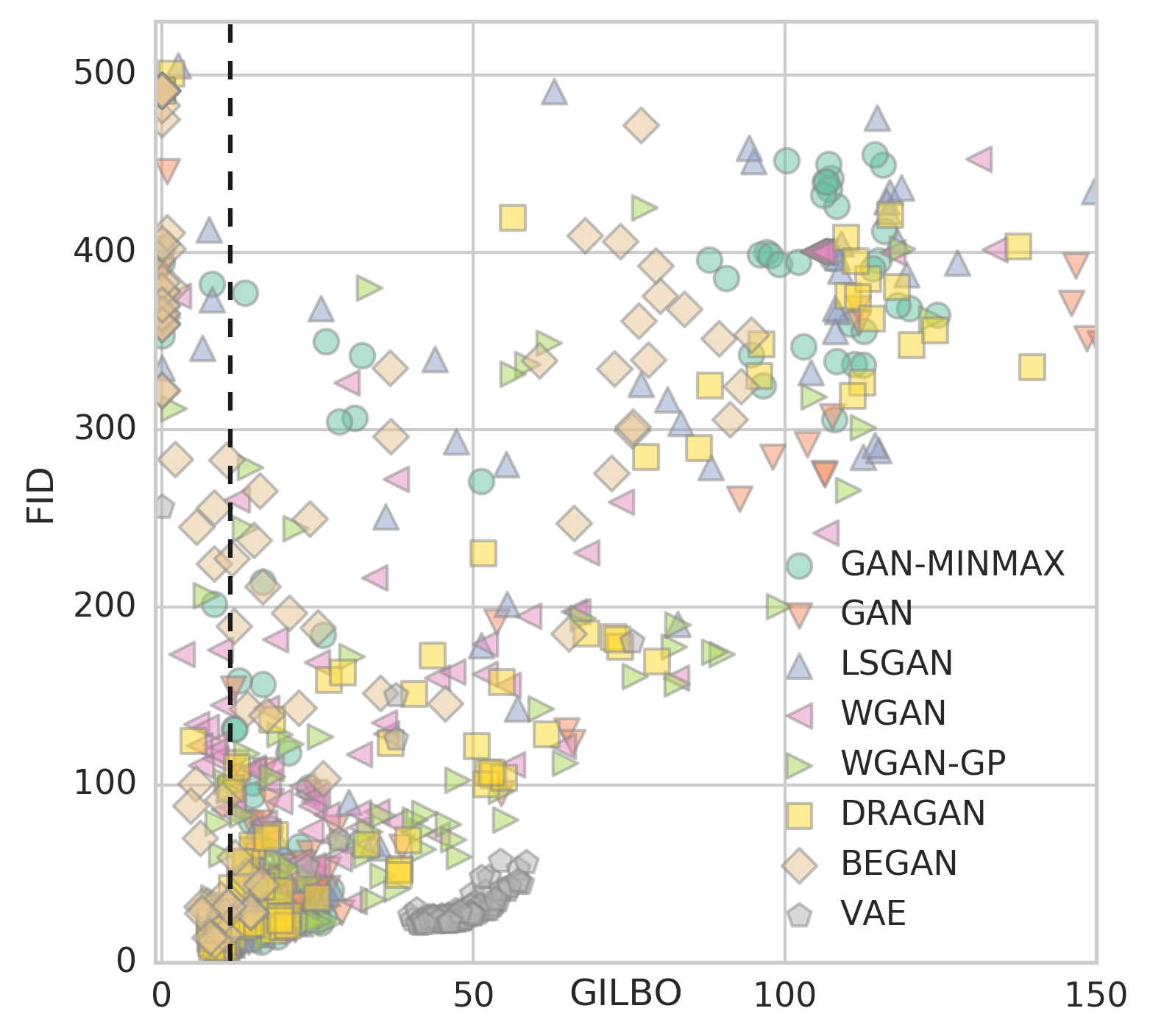}\label{fig:gilbofidscatter}}
	\caption{%
	(a) Recreation of Figure 5 (left) from \citet{gansequal}
	showing the distribution of \fid/ scores for each model on \textsc{mnist}. Points are jittered to give a sense of density.
	(b) The distribution of \gilbo/ scores.
	(c) \fid/ vs \gilbo/.
	}
	\label{fig:boxes}
\end{figure}

\begin{figure}[tbp]
	\centering
	\subfloat[\gilbo/ $\le \log C$]{\includegraphics[width=0.4\textwidth]{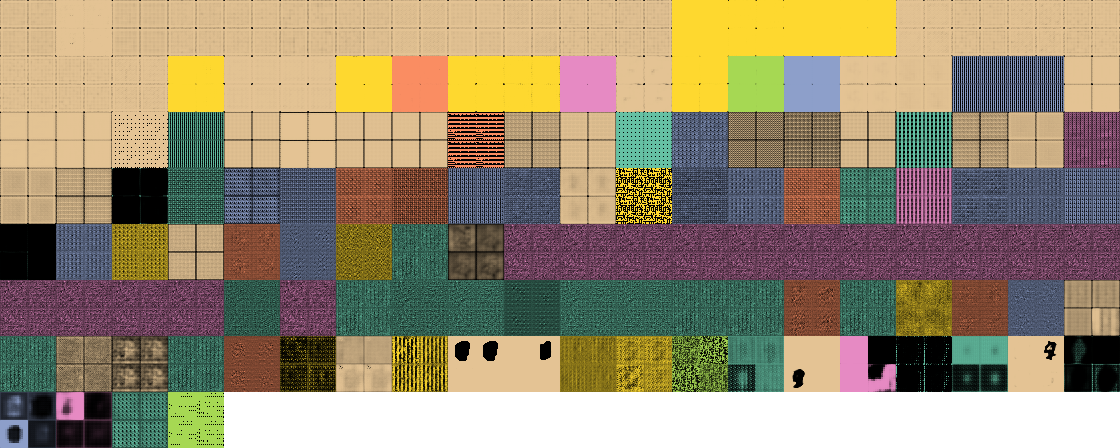}\label{fig:ln10}}
	\quad
	\subfloat[$\log C <$ \gilbo/ $\le \log N$]{\includegraphics[width=0.4\textwidth]{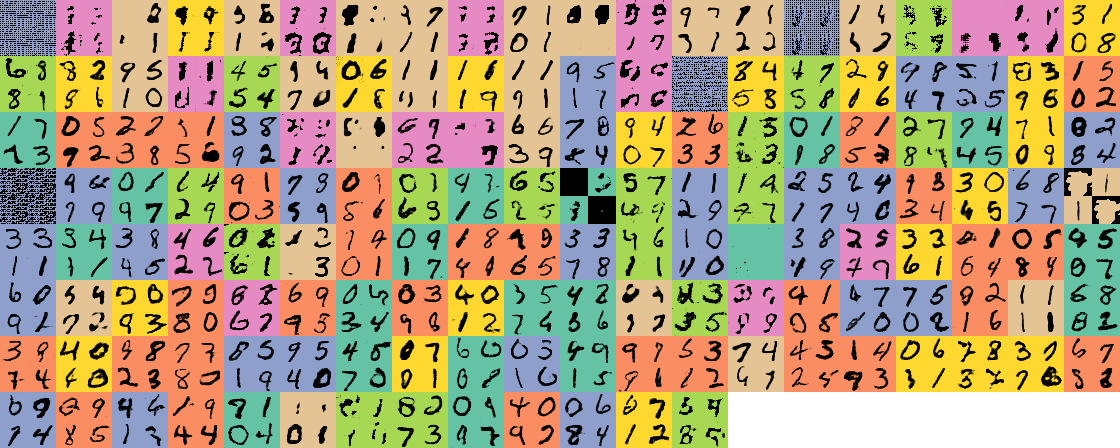}\label{fig:50k}} \\
	\subfloat[$\log N <$ \gilbo/ $\le 2 \log N$]{\includegraphics[width=0.4\textwidth]{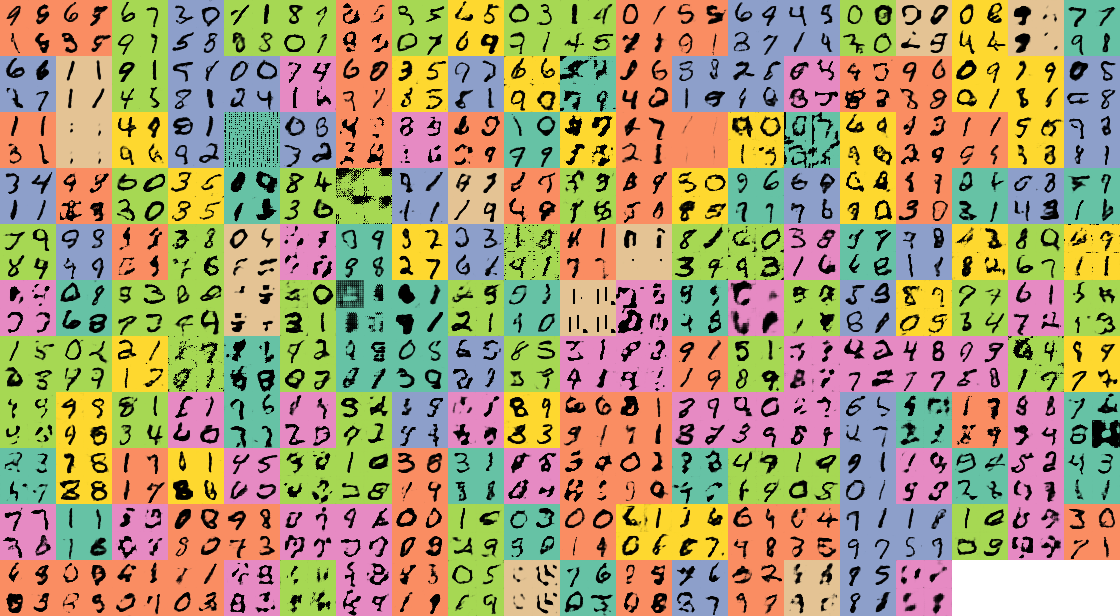}\label{fig:lt21}}
	\quad
	\subfloat[$2 \log N <$ \gilbo/ $\le 80.2 (\sim H(X))$]{\includegraphics[width=0.4\textwidth]{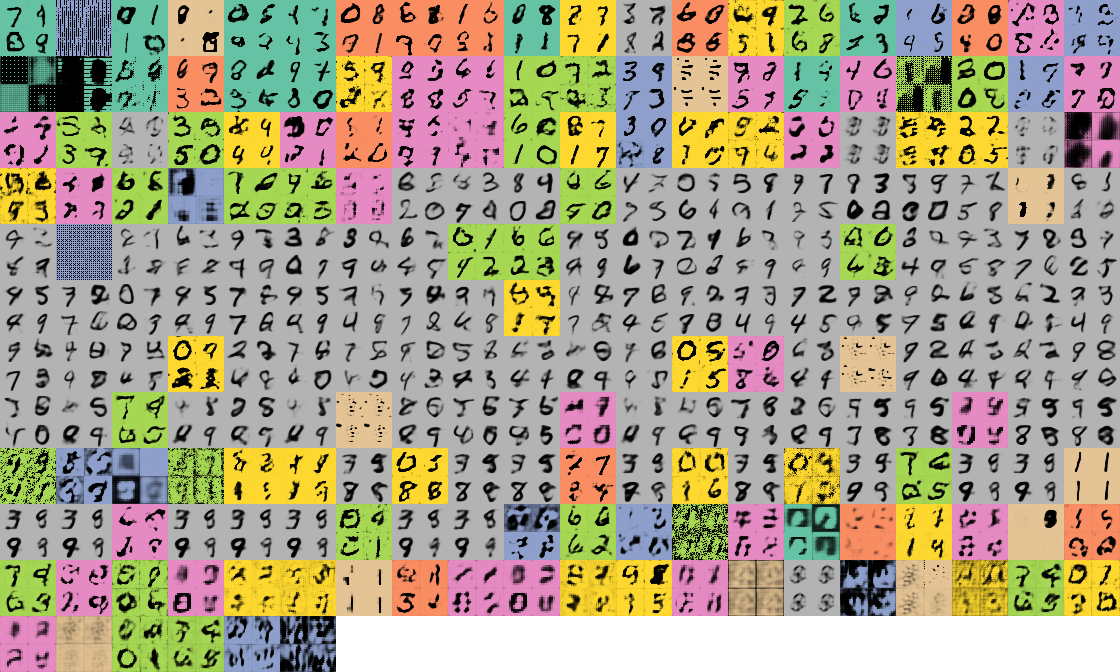}\label{fig:lt80}} \\
	\subfloat[$80.2 (\sim H(X)) < $ \gilbo/]{\includegraphics[width=0.4\textwidth]{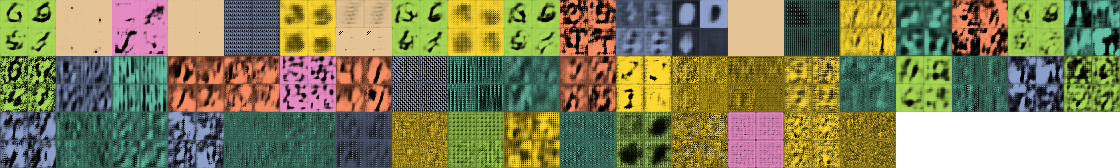}\label{fig:gt80}}
	\quad
	\subfloat[Legend]{\includegraphics[width=0.4\textwidth]{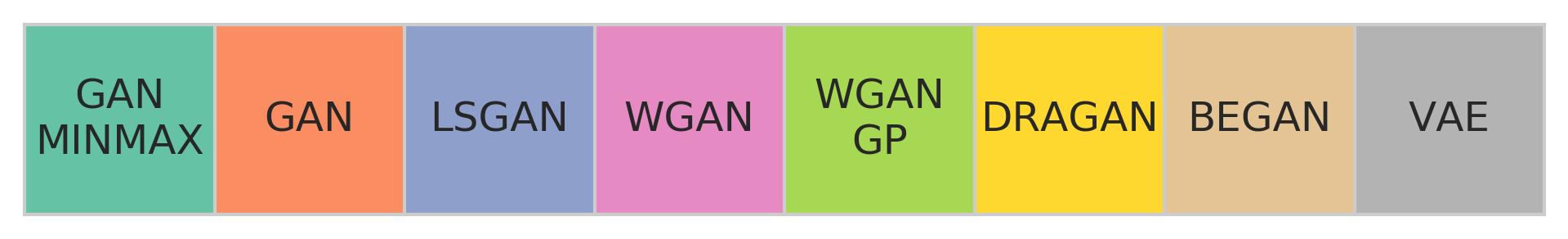}\label{fig:legend}}

	\caption{ Samples from all models sorted by increasing \gilbo/ in raster order and broken
	up into representative ranges. The background colors correspond to the
	model family (\Cref{fig:legend}).
	Note that all of the \vae/ samples are in (d), indicating that the \vae/s achieved a non-trivial amount of complexity.
	Also note that most of the \gan/s in (d) have poor sample quality, further underscoring the apparent difficulty these \gan/s
	have maintaining high visual quality without indications of training set memorization.
	}
	\label{fig:samps}
\end{figure}

In \Cref{fig:boxes} we show the distributions of \fid/ and \gilbo/ scores for all
800 models as well as their scatter plot for \textsc{mnist}. We can immediately see that
each of the \gan/ objectives collapse to \gilbo/ $ \sim 0$ for some
hyperparameter settings, but none of the \vae/s do.
In \Cref{fig:samps} we show generated samples from all of the models, split into
relevant regions.
A \gilbo/ near zero signals a failure of the model to make any use of its latent
space (\Cref{fig:ln10}).

The best performing models by \fid/ all sit at a
\gilbo/ $\sim 11$ nats.
An \textsc{mnist} model that simply memorized the
training set and partitioned the latent space into 50,000 unique outputs would
have a \gilbo/ of $\log 50,\!000 = 10.8$ nats, so the cluster around 11 nats
is suspicious. Since mutual information is invariant to any invertible transformation,
a model that partitioned the latent space into 50,000 bins, associated each with a training point
and then performed some random elastic transformation but with a magnitude low enough to not turn one
training point into another would still have a generative mutual information of 10.8 nats. Larger
elastic transformations that could confuse one training point for another would only act to lower
the generative information.
Among a large set of
hyperparameters and across 7 different \gan/ objectives, we notice a
conspicuous increase in \fid/ score as \gilbo/ moves away from $\sim 11$ nats to
either side.
This demonstrates the failure of these \gan/s to achieve a meaningful range of
complexities while maintaining visual quality. Most striking is the distinct separation
in \gilbo/s between \gan/s and \vae/s. These \gan/s learn less complex joint densities
than a vanilla \vae/ on \textsc{mnist} at the same \fid/ score.

\begin{figure}[tb]
	\centering
	\subfloat[\fid/]{\includegraphics[width=0.31\textwidth]{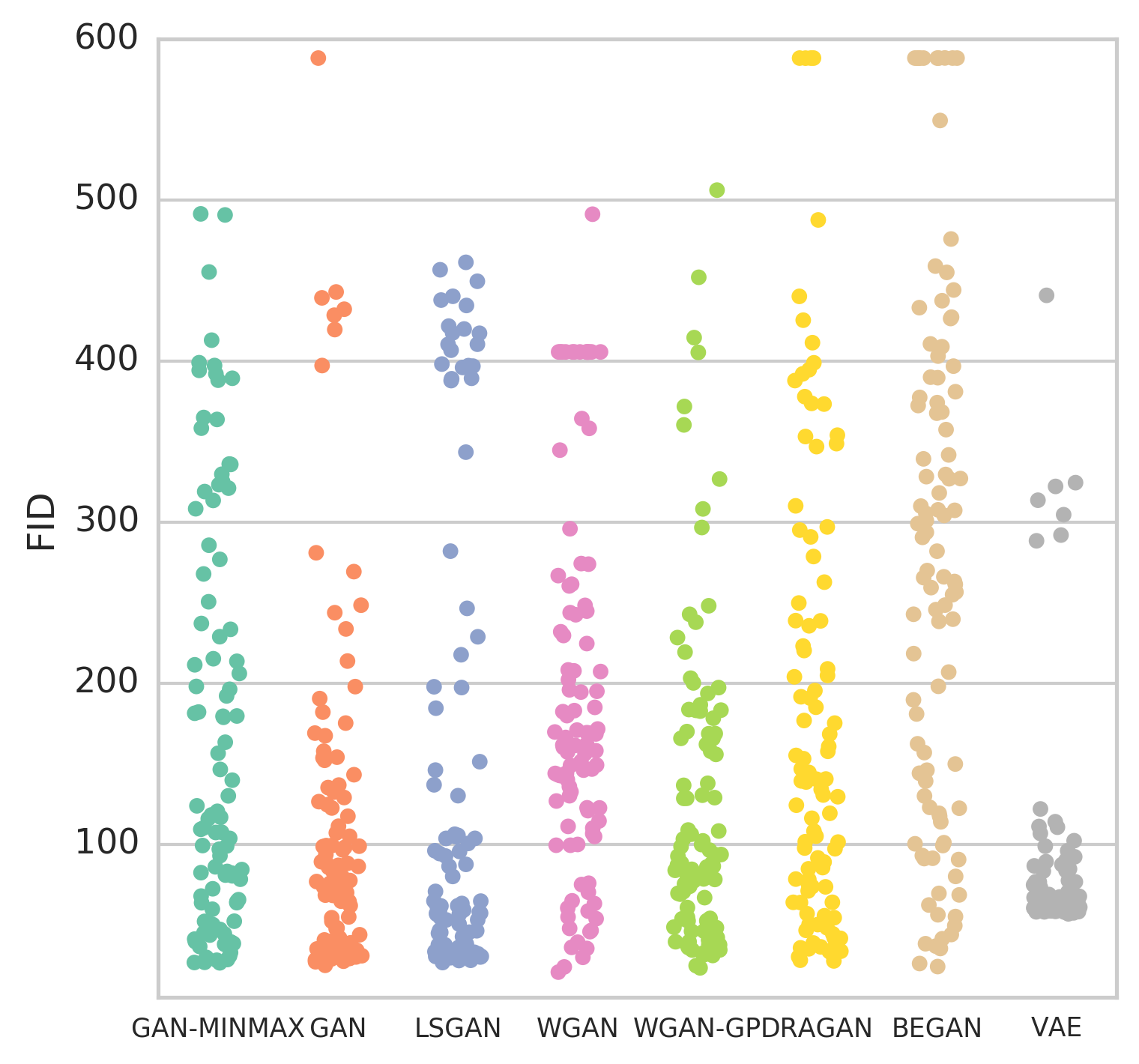}\label{fig:fashionfidbox}}
	\subfloat[\gilbo/]{\includegraphics[width=0.31\textwidth]{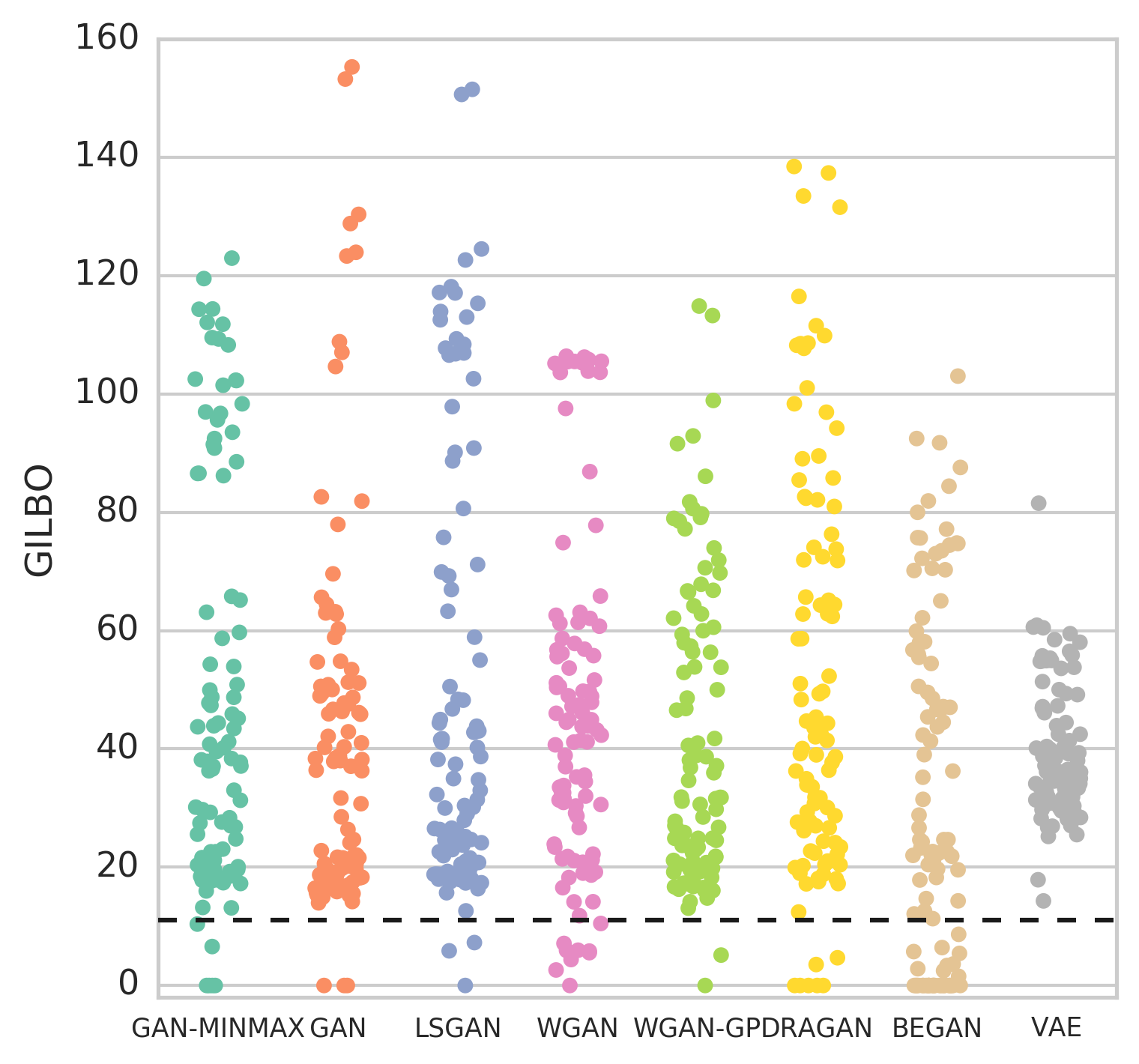}\label{fig:fashiongilbobox}}
	\subfloat[\gilbo/ vs \fid/]{\includegraphics[width=0.31\textwidth]{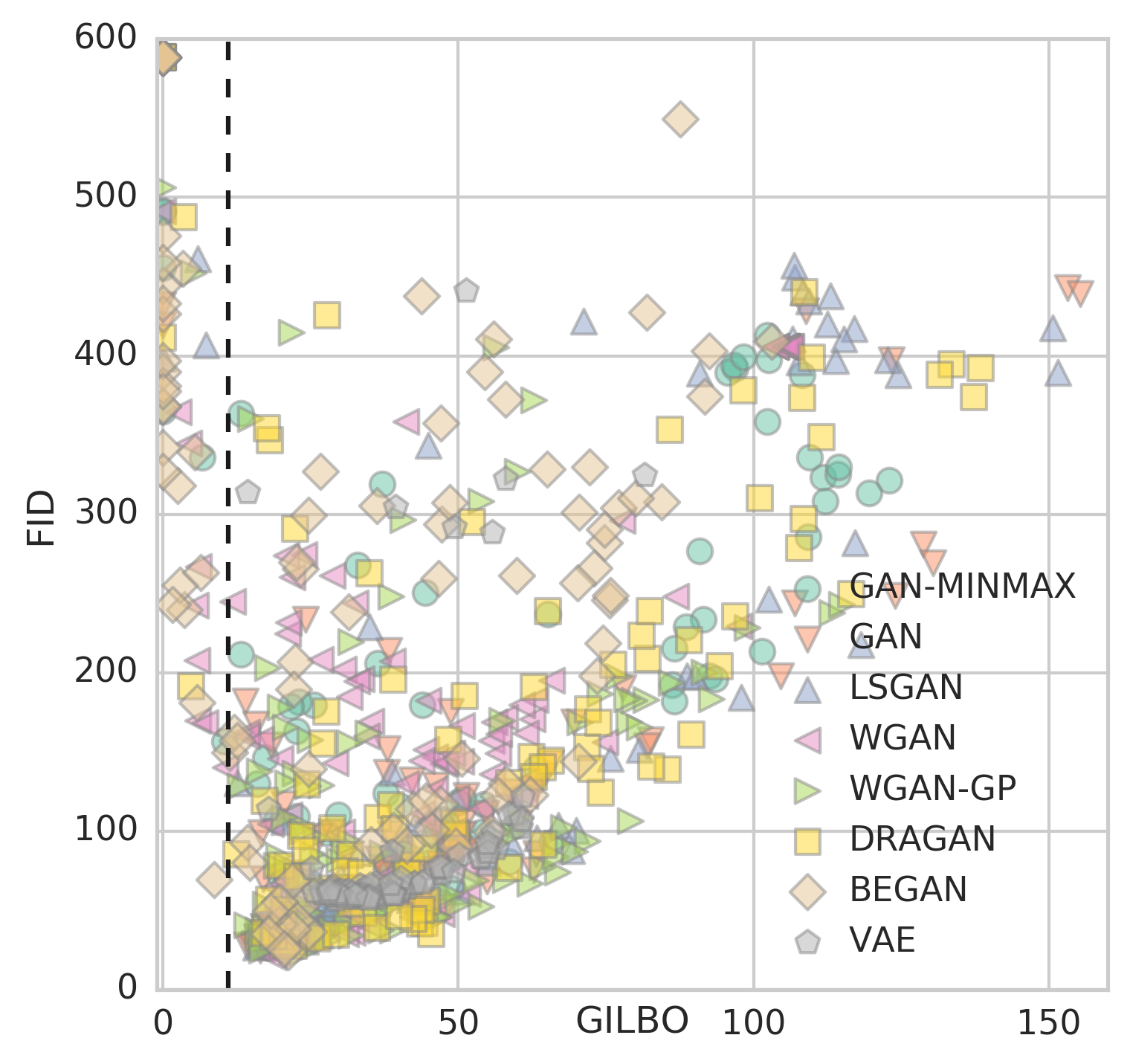}\label{fig:fashiongilbofidscatter}}
	\caption{A recreation of \Cref{fig:boxes} for the Fashion MNIST dataset.}
	\label{fig:fashionboxes}
\end{figure}

\begin{figure}[tb]
	\centering
	\subfloat[\fid/]{\includegraphics[width=0.31\textwidth]{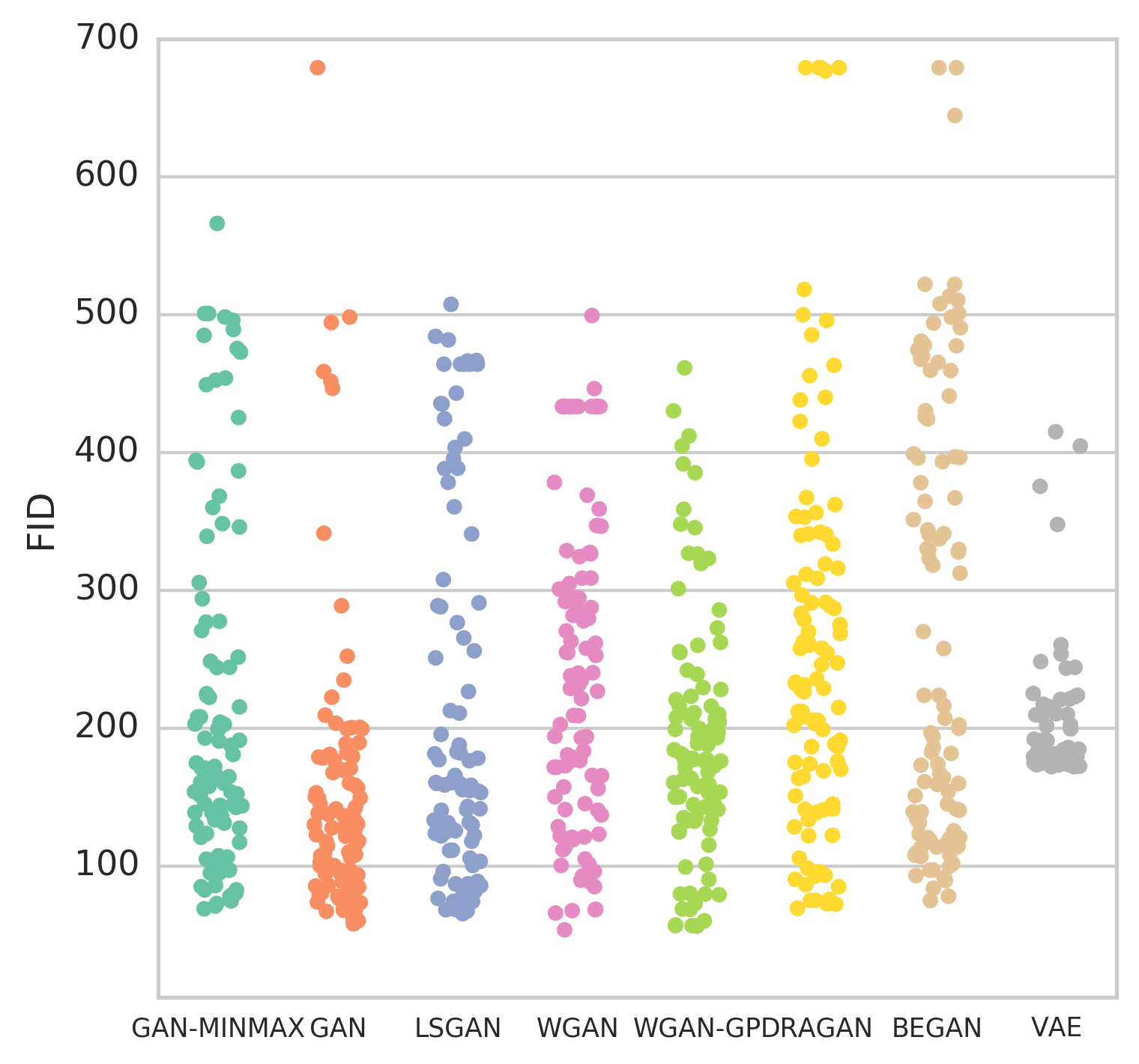}\label{fig:cifarfidbox}}
	\subfloat[\gilbo/]{\includegraphics[width=0.31\textwidth]{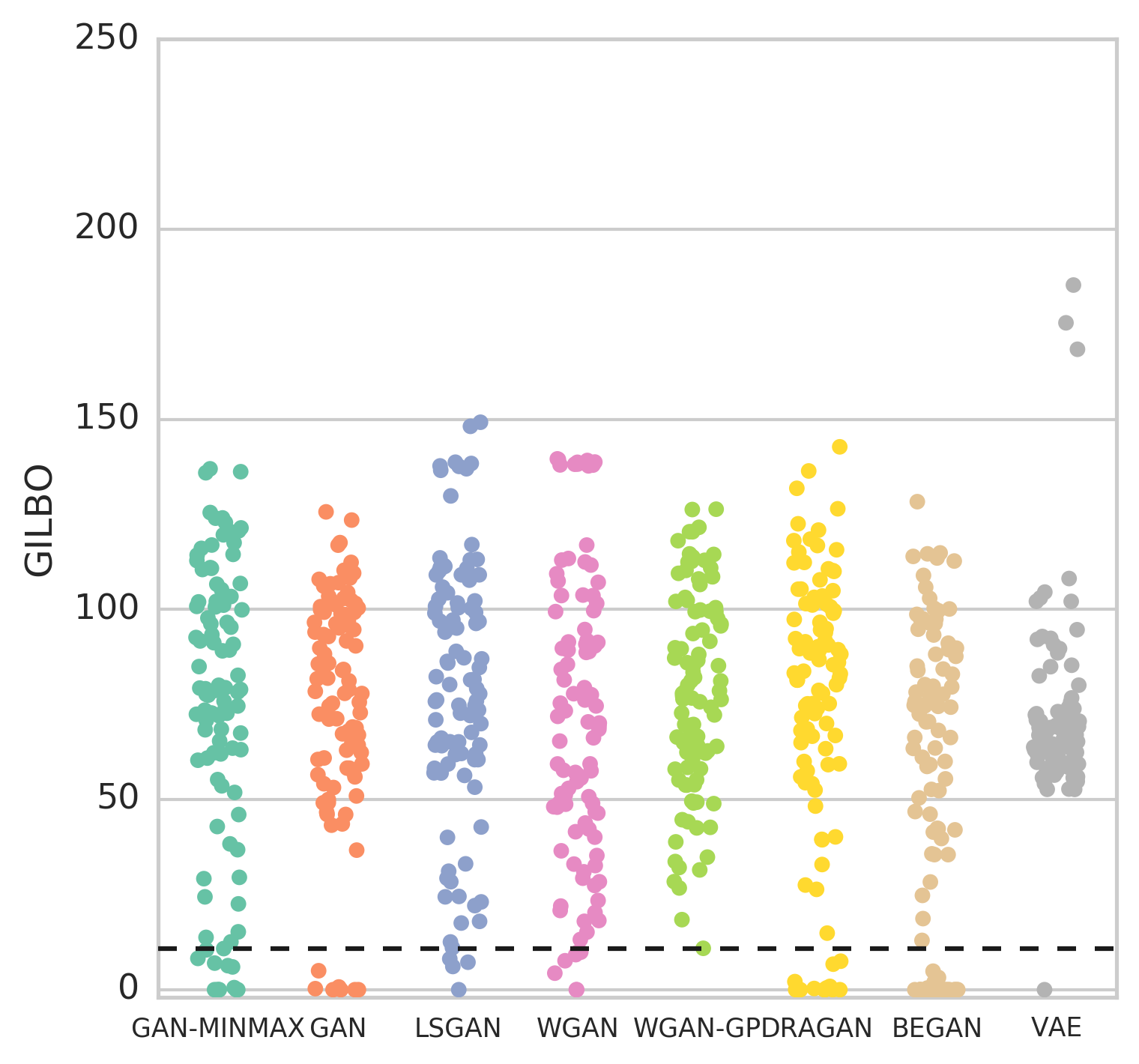}\label{fig:cifargilbobox}}
	\subfloat[\gilbo/ vs \fid/]{\includegraphics[width=0.31\textwidth]{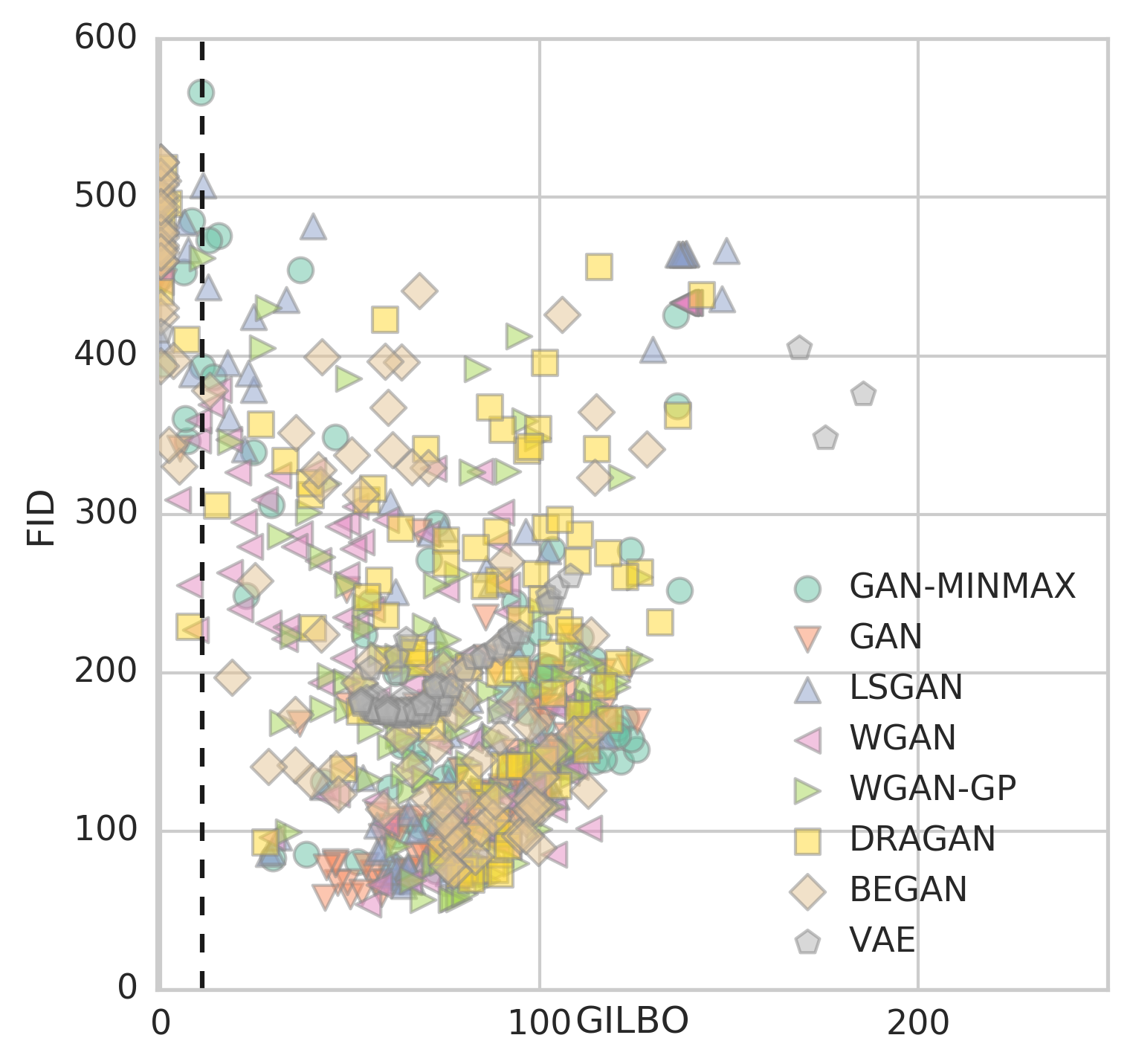}\label{fig:cifargilbofidscatter}}
	\caption{A recreation of \Cref{fig:boxes} for the CIFAR dataset.}
	\label{fig:cifarboxes}
\end{figure}

\begin{figure}[tb]
	\centering
	\subfloat[\fid/]{\includegraphics[width=0.31\textwidth]{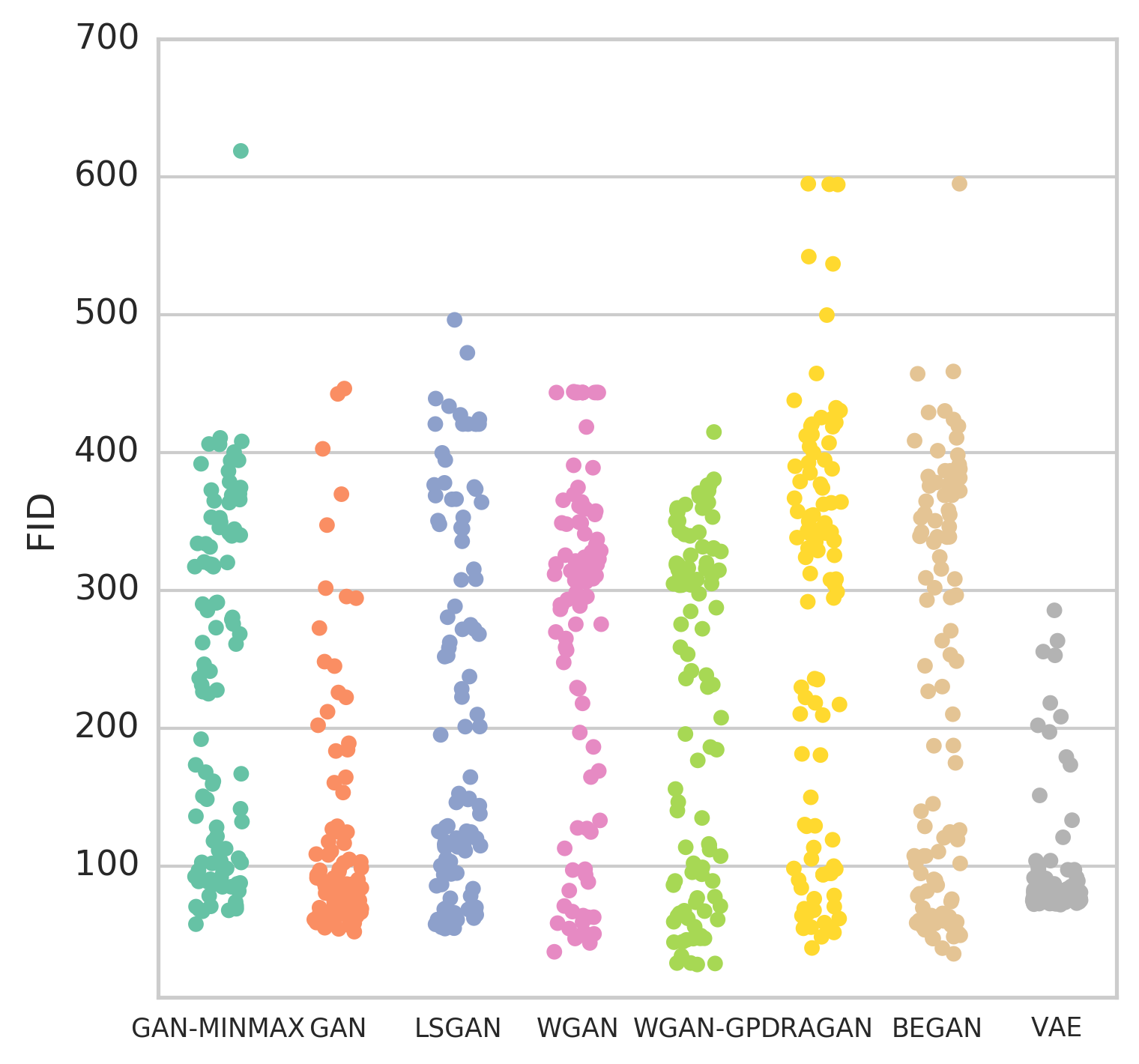}\label{fig:celebafidbox}}
	\subfloat[\gilbo/]{\includegraphics[width=0.31\textwidth]{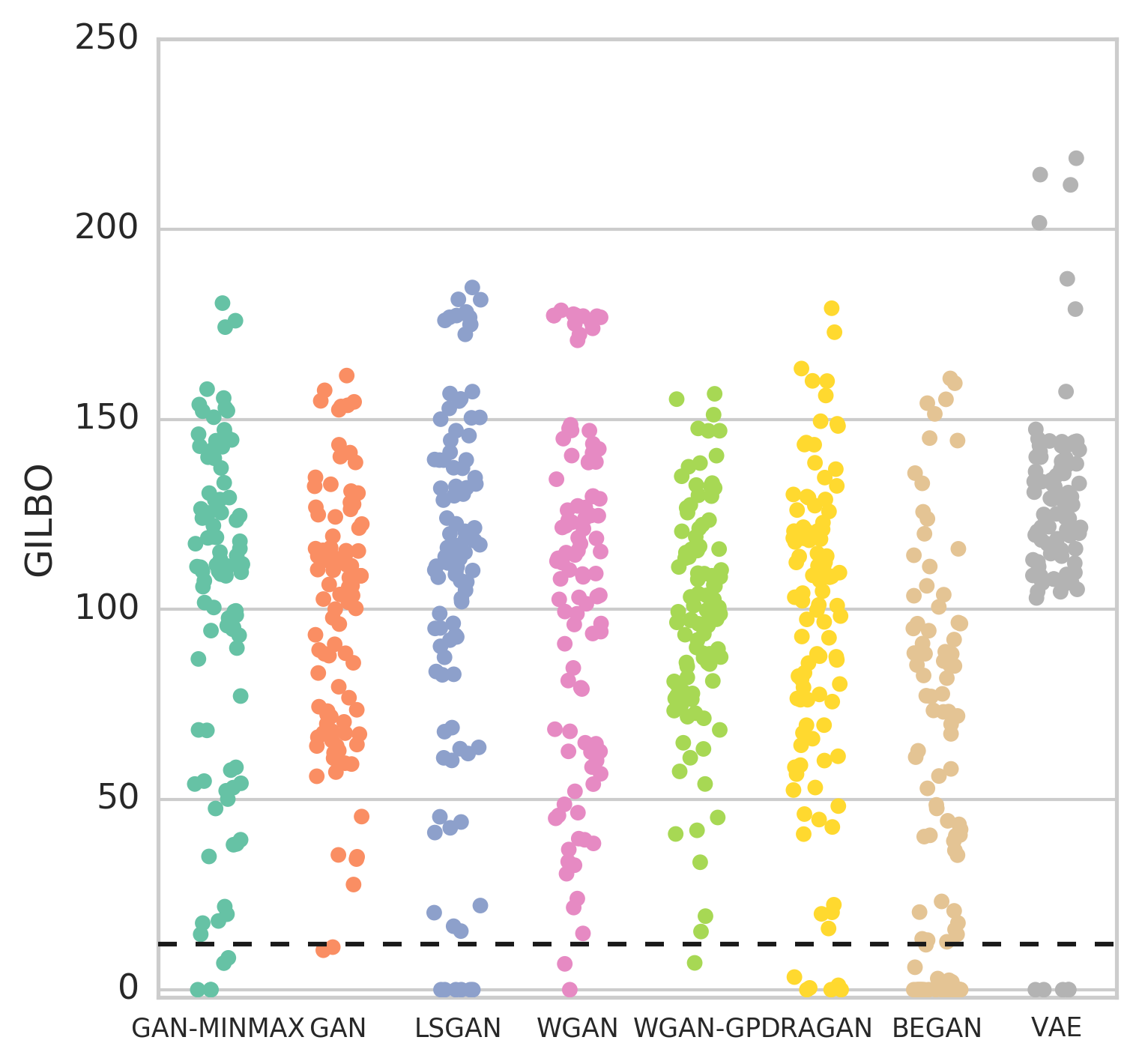}\label{fig:celebagilbobox}}
	\subfloat[\gilbo/ vs \fid/]{\includegraphics[width=0.31\textwidth]{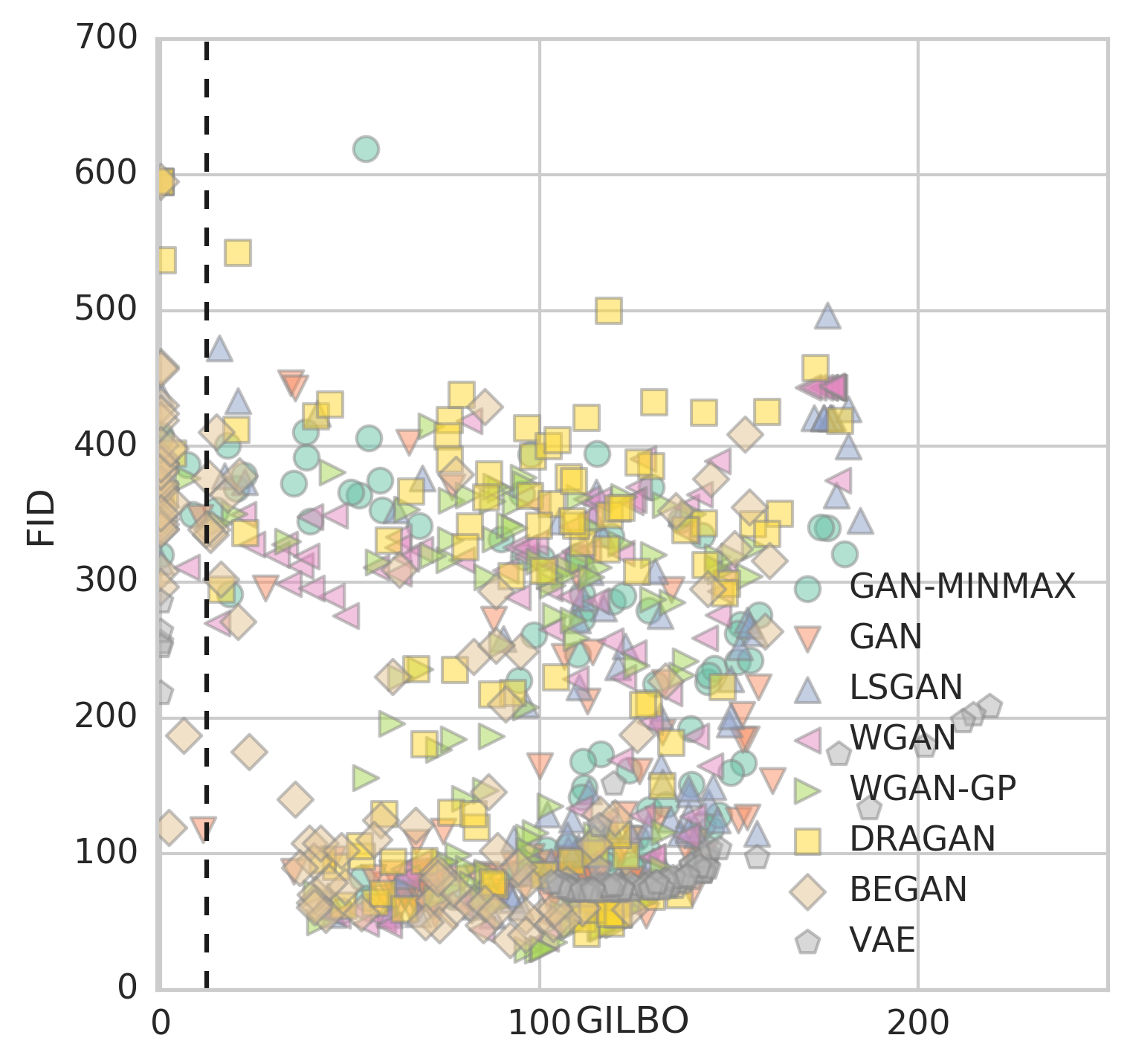}\label{fig:celebagilbofidscatter}}
	\caption{A recreation of \Cref{fig:boxes} for the CelebA dataset.}
	\label{fig:celebaboxes}
\end{figure}

\Cref{fig:fashionboxes,fig:cifarboxes,fig:celebaboxes} show the same
plots as in \Cref{fig:boxes} but for the Fashion\textsc{mnist},
\textsc{cifar}-10 and CelebA datasets respectively. The best performing models
as measured by \fid/ on Fashion\textsc{mnist} continue to have \gilbo/s near
$\log N$. However, on the more complex \textsc{cifar}-10 and CelebA datasets
we see nontrivial variation in the complexities of the trained \gan/s with competitive \fid/.
On these more complex datasets, the visual performance (e.g. \Cref{fig:consistency}) of
the models leaves much to be desired. We speculate that the models' inability to acheive high visual quality
is due to insufficient model capacity for the dataset.

}%

	\section{Discussion}

\subsection{Reproducibility}

\begin{figure}[tbp]
	\centering
	\subfloat[\gilbo/ $\sim 41.1 \pm 0.8$ nats]{\includegraphics[width=0.31\textwidth]{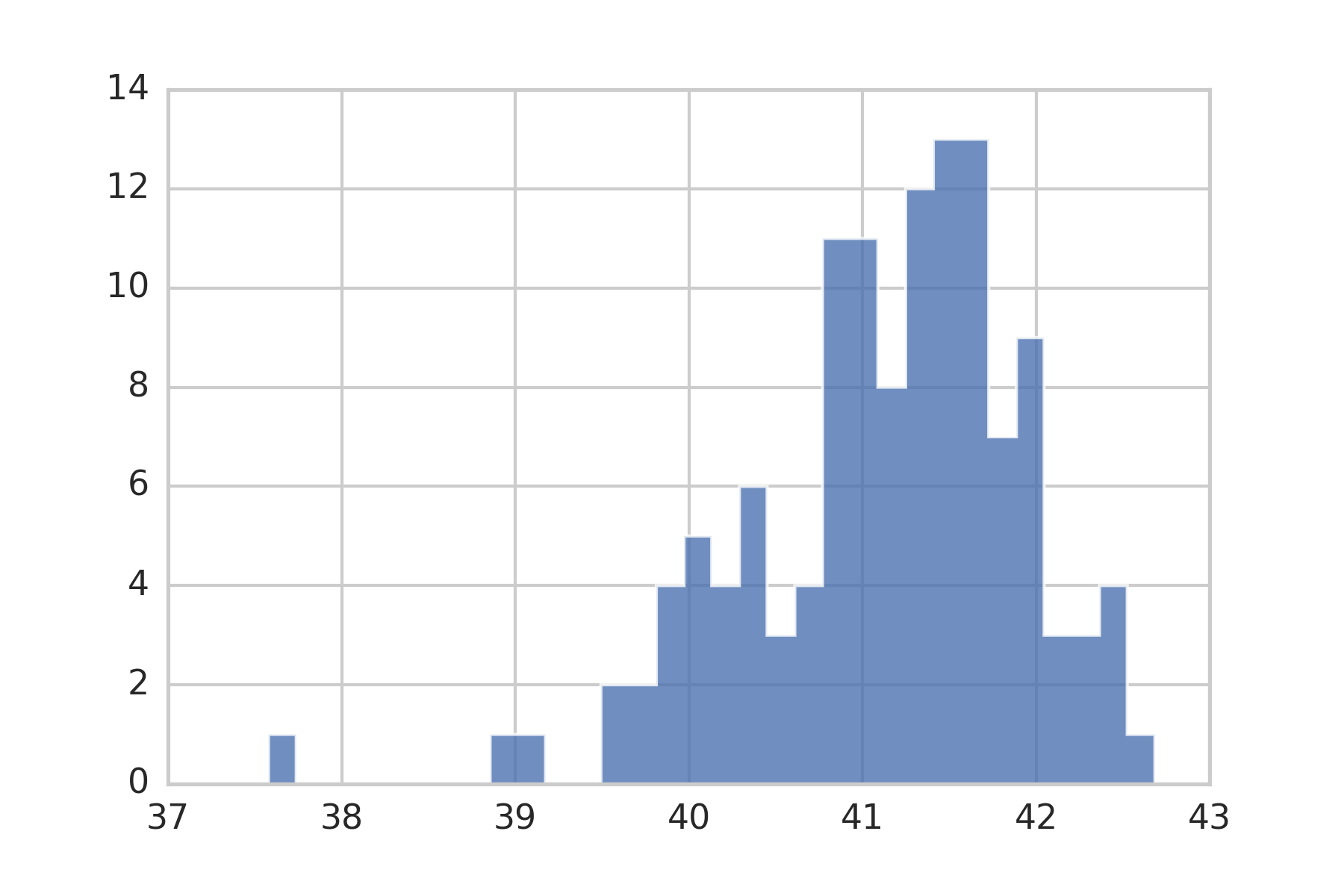}\label{fig:repsmall}}
	\quad
	\subfloat[\gilbo/ $\sim 69.5 \pm 0.9$ nats]{\includegraphics[width=0.31\textwidth]{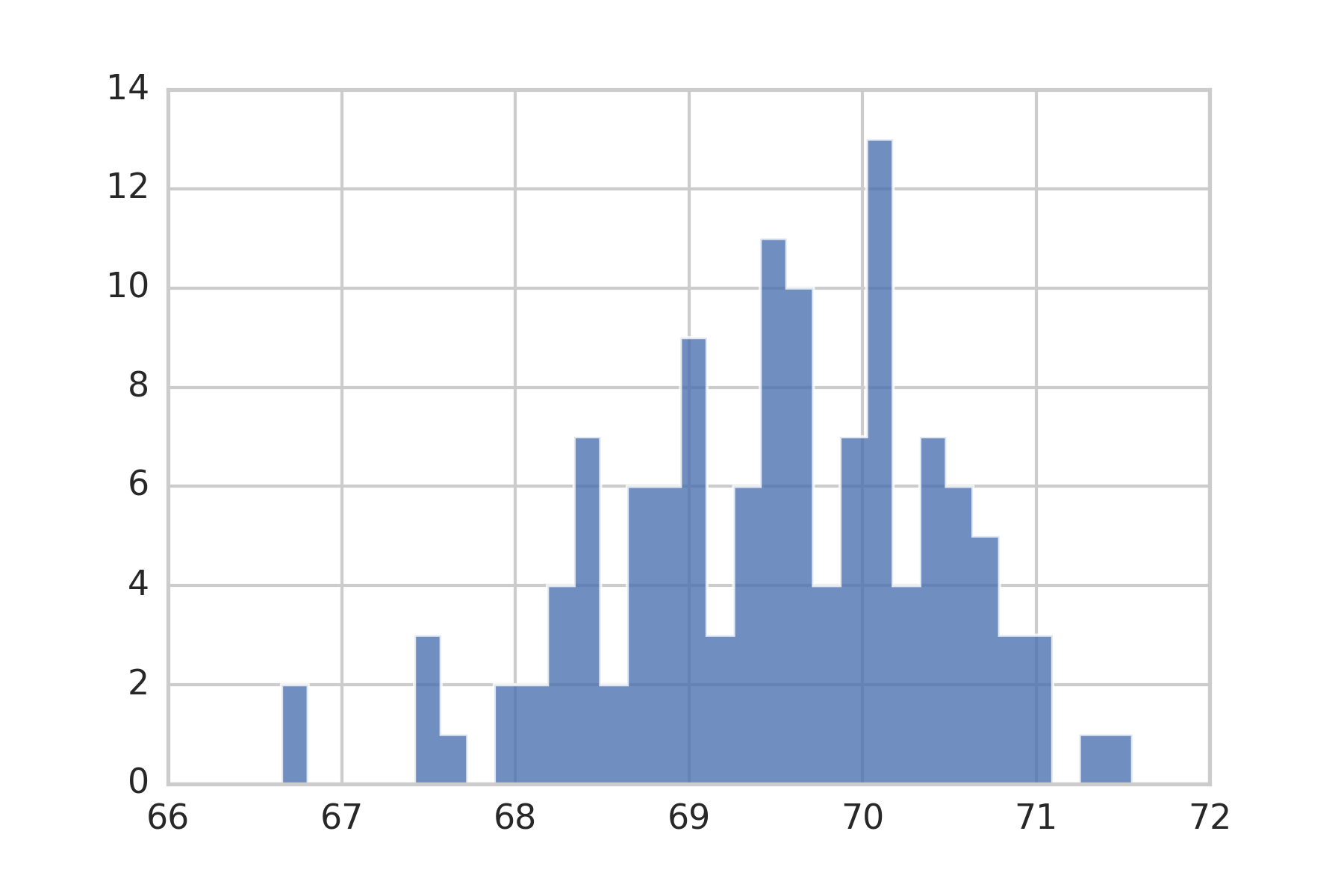}\label{fig:repmedium}}
	\quad
	\subfloat[\gilbo/ $\sim 104 \pm 1$ nats]{\includegraphics[width=0.31\textwidth]{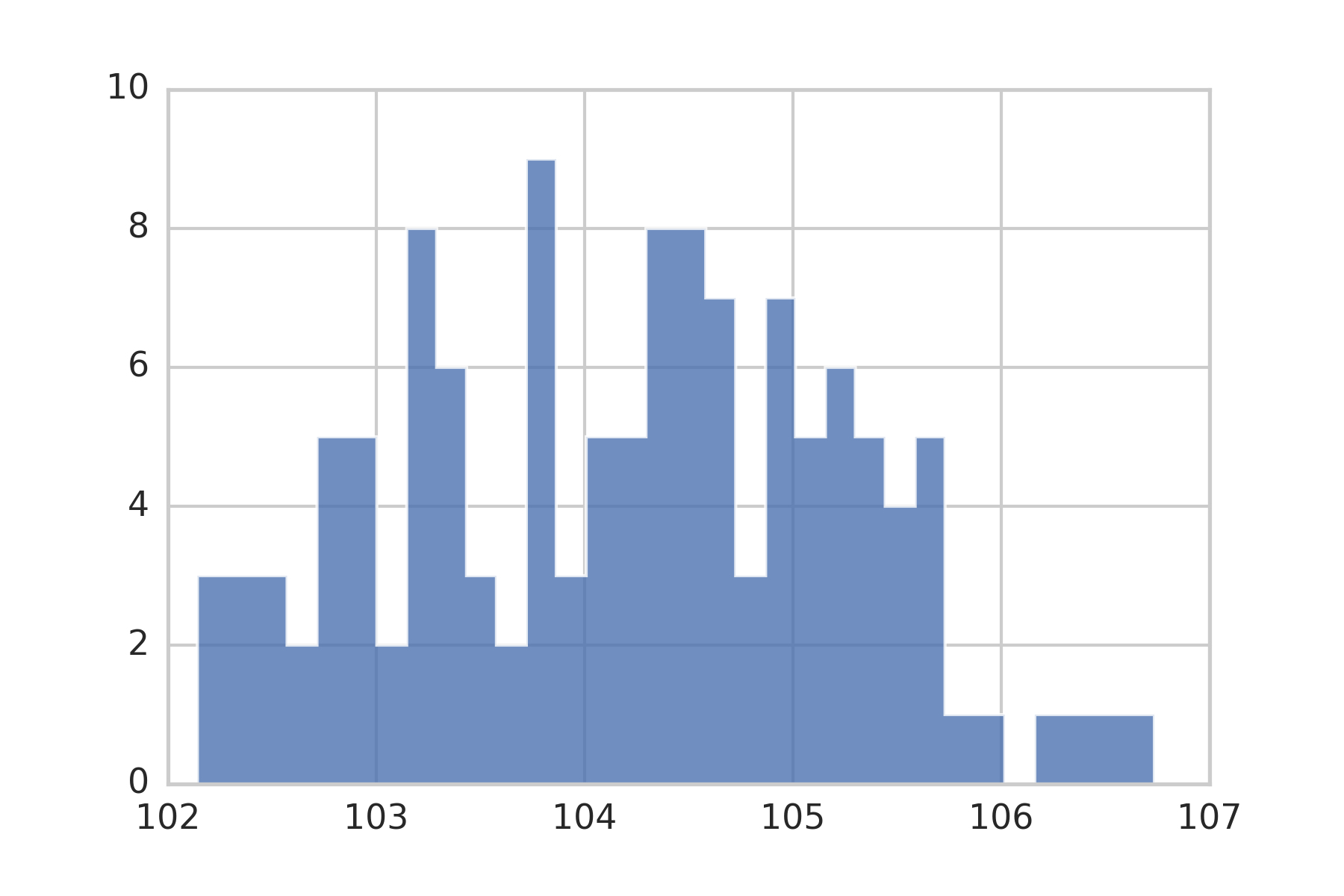}\label{fig:replarge}}
	\caption{
		Measure of the reproducibility of the \gilbo/ for
		the three models visualized in \Cref{fig:consistency}.
		For each model we independently measured the \gilbo/ 128 times.
	}
	\label{fig:reproduce}
\end{figure}

While the \gilbo/ is a valid lower bound regardless of the accuracy of the
learned encoder, its utility as a metric naturally requires it to be comparable
across models. The first worry is whether it is reproducible in its values.  To
address this, in \Cref{fig:reproduce} we show the result of 128 different
training runs to independently compute the \gilbo/ for three models on CelebA.
In each case the error in the
measurement was below 2\% of the mean \gilbo/ and much smaller in variation than the
variations between models, suggesting comparisons between models are valid if
we use the same encoder architecture ($e(z|x)$) for each.

\subsection{Tightness}
\label{sec:sbc}

Another concern would be whether the learned variational encoder was a good
match to the true posterior of the generative model ($e(z|x) \sim p(z|x)$).
Perhaps the model with a measured \gilbo/ of 41 nats simply had a harder
to capture $p(z|x)$ than the $\gilbo/ \sim 104$ nat model.  Even if the values
were reproducible between runs, maybe there is a systemic bias in the approximation
that differs between different models.

\begin{figure}[tbp]
	\centering
	\subfloat[\gilbo/ $\sim$ 41 nats]{\includegraphics[width=0.31\textwidth]{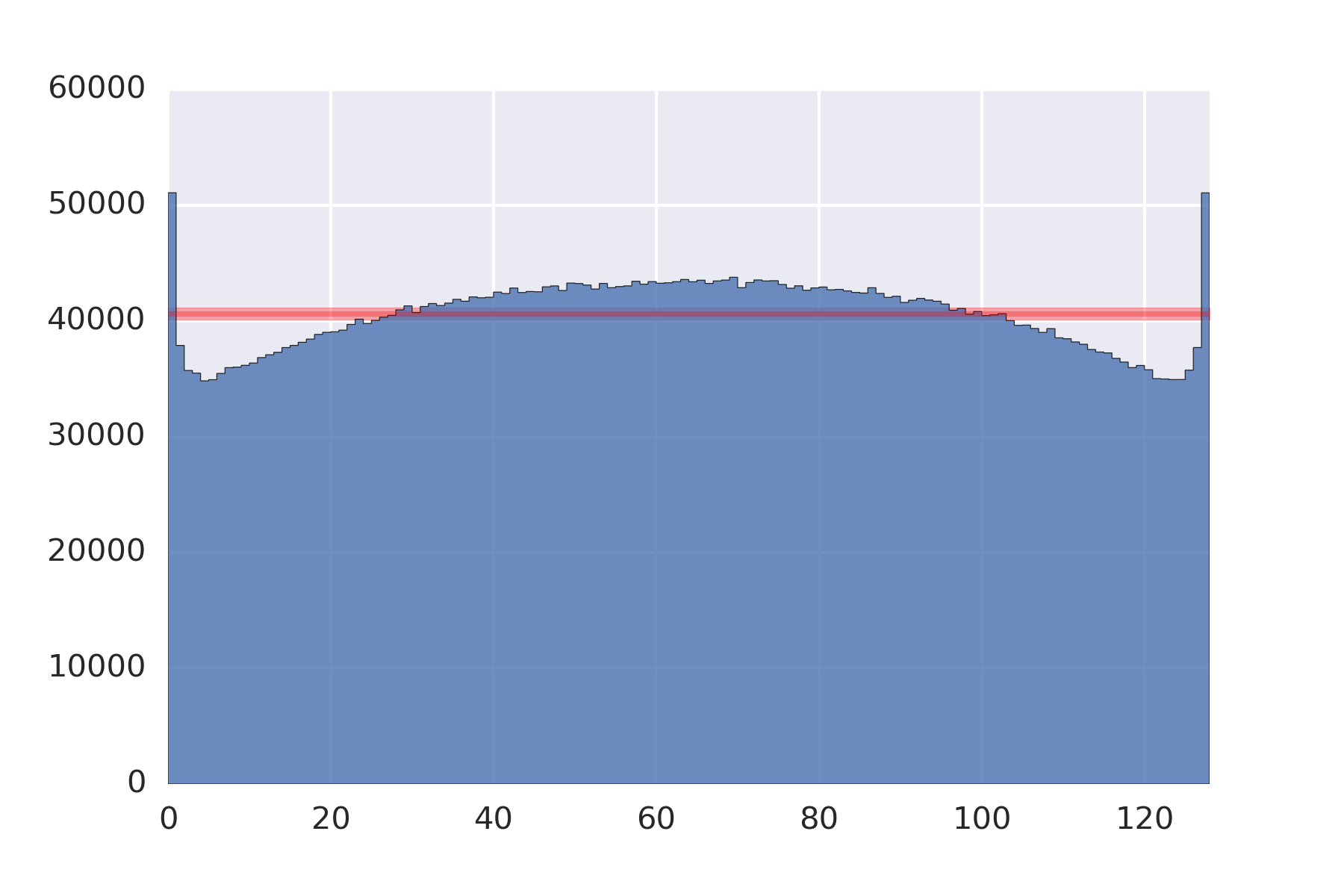}\label{fig:sbcsmall}}
	\quad
	\subfloat[\gilbo/ $\sim$ 70 nats]{\includegraphics[width=0.31\textwidth]{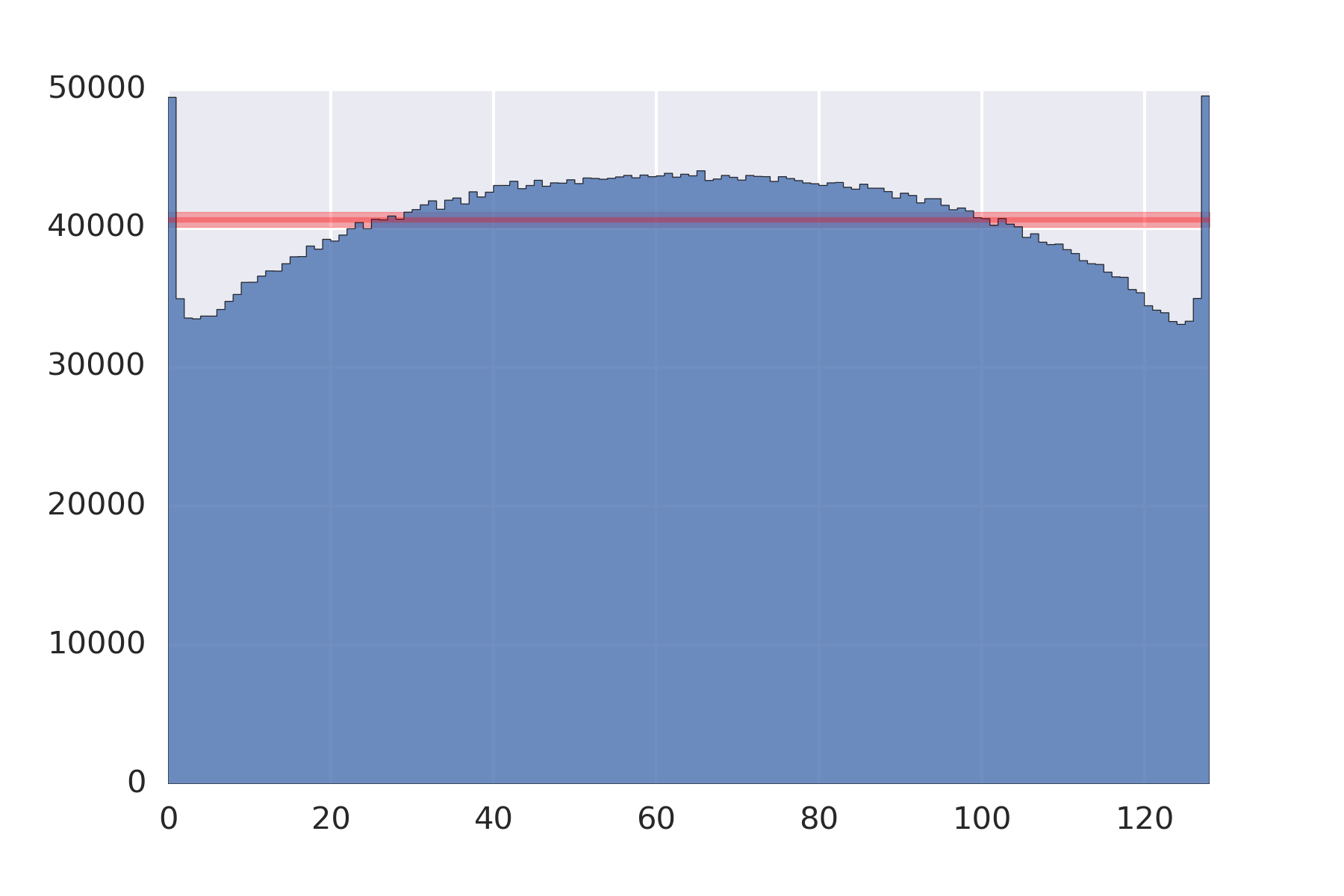}\label{fig:sbcmedium}}
	\quad
	\subfloat[\gilbo/ $\sim$ 104 nats]{\includegraphics[width=0.31\textwidth]{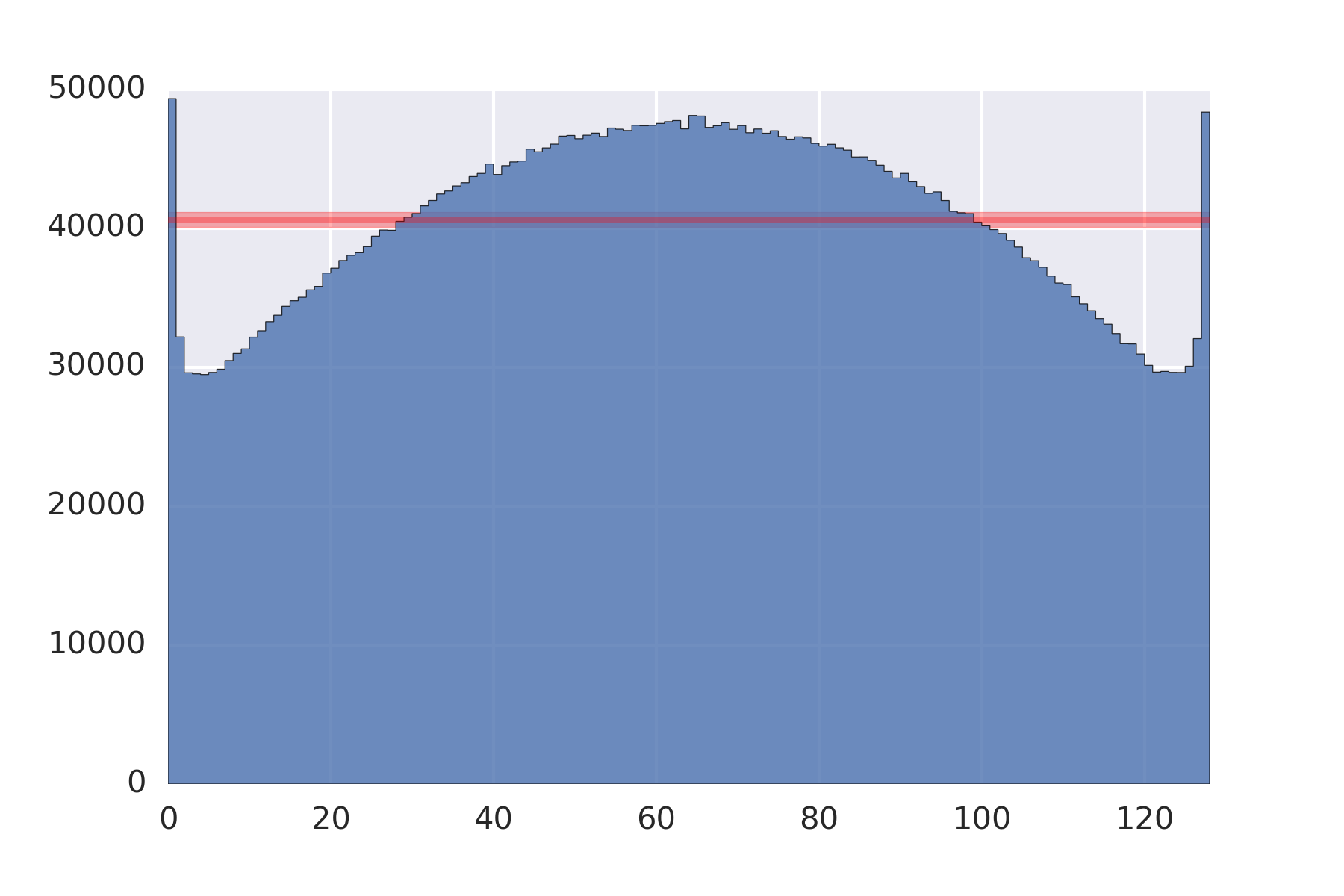}\label{fig:sbclarge}}
	\caption{
		Simulation-based calibration~\citep{sbc} of the variational
		encoder for the same three models as in \Cref{fig:reproduce}.
		Shown are histograms of the ranking statistic for how many of
		128 samples from the encoder are less than the true $z$ used to
		generate the figure, aggregated over the 64 dimensional latent
		vector for 1270 batches of 64 samples each. Shown in red is the
		99\% confidence interval for a uniform distribution, the
		expected result if $e(z|x)$ was the true $p(z|x)$. The
		systematic $\cap$-shape denotes overdispersion in the
		approximation.
	}
	\label{fig:sbc}
\end{figure}

To test this, we used the \emph{Simulation-Based Calibration} (\textsc{SBC}) technique
of \citet{sbc}.  If one were to implement a cycle, wherein a single draw from
the prior $z' \sim p(z)$ is decoded into an image $x' \sim p(x|z')$ and then
inverted back to its corresponding latent $z_i \sim p(z|x')$, the rank
statistic $\sum_i \mathbb{I}\left[z_i < z'\right]$ should be uniformly
distributed. Replacing the true $p(z|x')$ with the approximate $e(z|x)$ gives a
visual test for the accuracy of the approximation. \Cref{fig:sbc} shows a
histogram of the rank statistic for 128 draws from $e(z|x)$ for each of 1270
batches of 64 elements each drawn from the 64 dimensional prior $p(z)$ for the
same three \gan/s as in \Cref{fig:reproduce}.  The red line denotes the 99\%
confidence interval for the corresponding uniform distribution.  All three
\gan/s show a systematic $\cap$-shaped distribution denoting overdispersion in
$e(z|x)$ relative to the true $p(z|x)$.  This is to be expected from a
variational approximation, but importantly the degree of mismatch seems to
correlate with the scores, not anticorrelate. It is likely that the 41 nat \gilbo/ is
a more accurate lower bound than the 103 nat \gilbo/.  This further reinforces
the utility of the \gilbo/ for cross-model comparisons.

\subsection{Precision of the \Gilbo/}
\label{sec:precision}

While comparisons between models seem well-motivated, the \textsc{SBC} results in~\Cref{sec:sbc}
highlight some mismatch in the variational approximation. How well can we
trust the absolute numbers computed by the \gilbo/? While they are guaranteed
to be valid lower bounds, how tight are those bounds?

To answer these questions, note that the \gilbo/ is a valid lower bound even if we learn
separate per-instance variational encoders.
Here we replicate the results of \citet{precise} and attempt to learn the precise $z$ that gave rise to an image
by minimizing the $L^2$ distance between the produced image and the target
($\left| x - g(z) \right|^2$).
We can then define a distribution centered on $z$ and adjust the magnitude of the variance to get the best \gilbo/ possible.
In other words, by minimizing the $L^2$ distance between an image $x$ sampled from the generative model and some \textit{other} $x'$ sampled from the same model, we can directly recover some $z'$ equivalent to the $z$ that generated $x$.
We can then do a simple optimization to find the variance that maximizes the \gilbo/, allowing us to compute a very tight \gilbo/ in a very computationally-expensive manner.

Doing this procedure on the same three models as in \Cref{fig:reproduce,fig:sbc}
gives (87, 111, 155) nats respectfully for the (41, 70, 104) \gilbo/
models, when trained for 150k steps to minimize the $L^2$ distance. These
approximations are also valid lower bounds, and demonstrate that our amortized
\gilbo/ calculations above might be off by as much as a factor of 2 in their
values from the true generative information, but again highlights that the
comparisons between different models appear to be real.
Also note that these per-image bounds are finite.
We discuss the finiteness of the generative information in more detail in~\Cref{sec:finite}.

Naturally, learning a single parametric amortized variational encoder is much
less computationally expensive than doing an independent optimization for each
image, and still seems to allow for comparative measurements.
However, we caution against directly comparing \gilbo/ scores from different encoder
architectures or optimization procedures.
Fair comparison between models requires holding the encoder architecture and training procedure fixed.

\subsection{Consistency}

\begin{figure}[!t]
	\centering
	\subfloat[\gilbo/ $\sim$ 41 nats]{\includegraphics[width=0.31\textwidth]{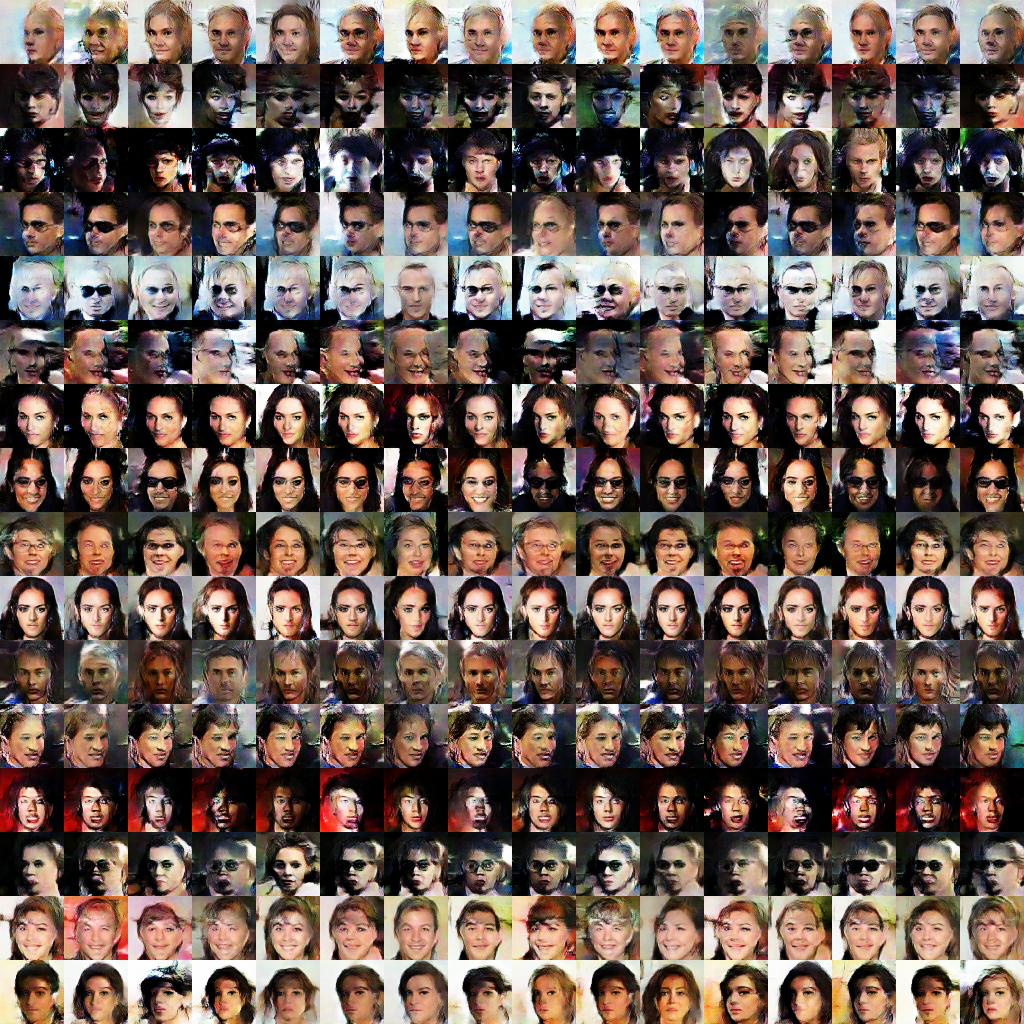}\label{fig:small}}
	\quad
	\subfloat[\gilbo/ $\sim$ 70 nats]{\includegraphics[width=0.31\textwidth]{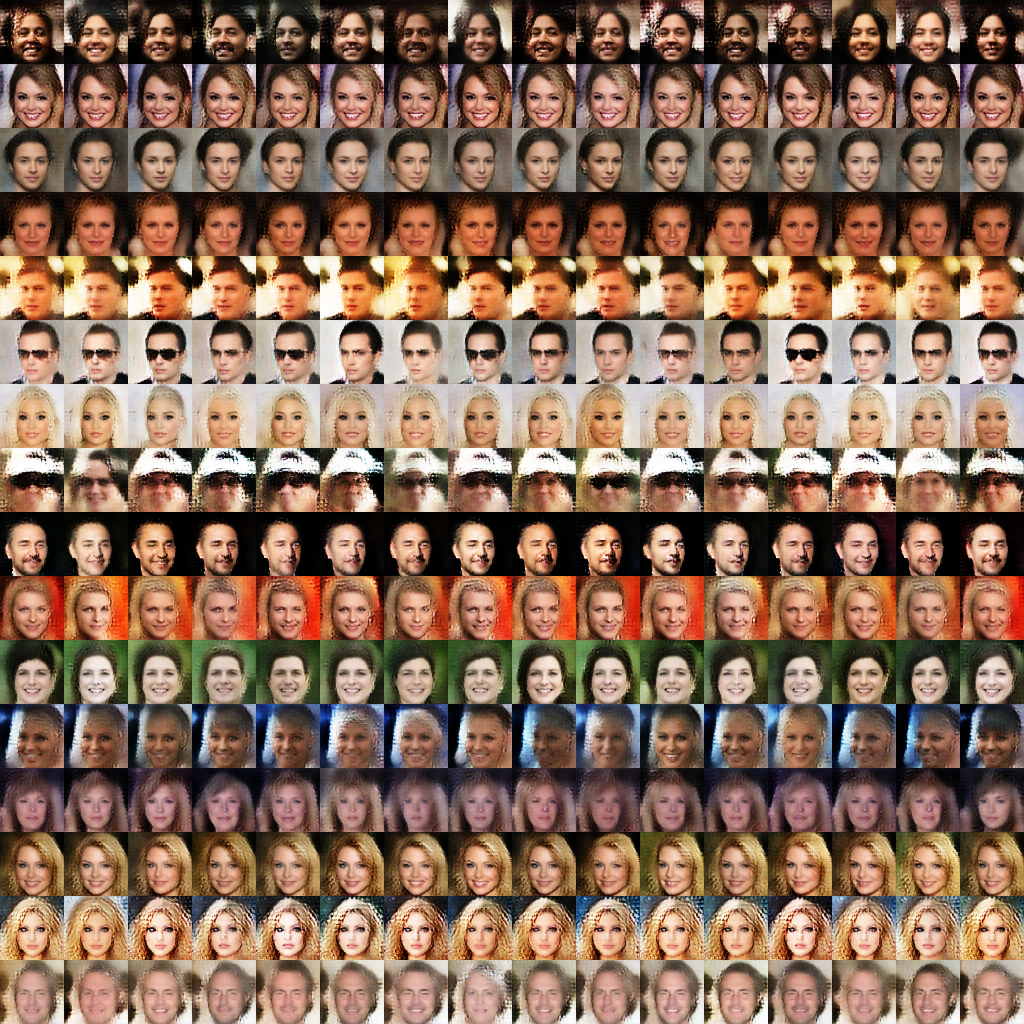}\label{fig:medium}}
	\quad
	\subfloat[\gilbo/ $\sim$ 104 nats]{\includegraphics[width=0.31\textwidth]{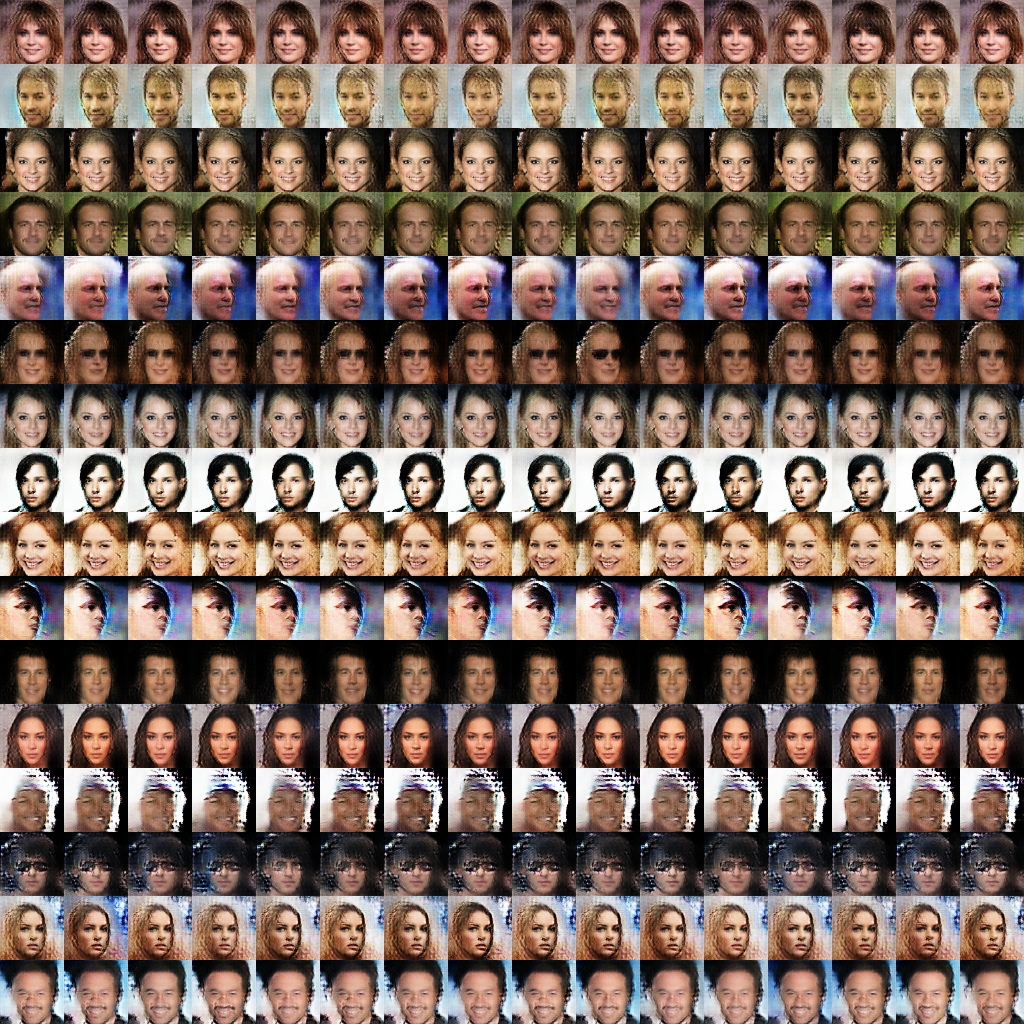}\label{fig:large}}
	\caption{Visual demonstration of consistency.  The left-most column of each image shows a sampled image from the \gan/.  The next 15 columns
	show images generated from 15 independent samples of the latent code suggested for the left-most image by the trained
	encoder used in estimating the \gilbo/. All three of these \gan/s had an \fid/ of 49 on CelebA, but have qualitatively different
	behavior.}
	\label{fig:consistency}
\end{figure}

The \gilbo/ offers a signal distinct from data-based metrics
like \fid/.  In \Cref{fig:consistency}, we visually demonstrate
the nature of the retained information for the same three models as above in \Cref{fig:reproduce,fig:sbc}.
All three checkpoints for CelebA have the same \fid/ score of 49,
making them
each competitive amongst the \gan/s studied; however, they have \gilbo/ values that
span a range of 63 nats (91 bits), which indicates a massive difference in model complexity.
In each figure, the left-most column shows a set of
independent generated samples from the \gan/. Each of these generated images
are then sent through the variational encoder $e(z|x)$ from which 15
independent samples of the corresponding $z$ are drawn.  These latent codes are
then sent back through the \gan/'s generator to form the remaining 15 columns.

The images in~\Cref{fig:consistency} show the type of information that is retained in the mapping
from image to latent and back to image space. On the right in \Cref{fig:large} with a
\gilbo/ of 104 nats, practically all of the human-perceptible information is retained
by doing this cycle. In contrast, on the left in \Cref{fig:small} with a \gilbo/ of only
41 nats, there is a good degree of variation in the synthesized images, although they
generally retain the overall gross attributes of the faces. In the middle, at 70 nats, the variation in
the synthesized images is small, but noticeable, such as the sunglasses that appear and disappear 6 rows from the top.

\subsection{Overfitting of the \Gilbo/ Encoder}

Since the \gilbo/ is trained on generated samples, the dataset is limited only by the number of unique samples the generative model can produce.
Consequently, it should not be possible for the encoder, $e(z|x)$, to overfit to the training data.
Regardless, when we actually evaluate the \gilbo/, it is always on newly generated data.

Likewise, given that the \gilbo/ is trained on the ``true'' generative model $p(z)p(x|z)$, we do not expect regularization to be necessary.
The encoders we trained are unregularized.
However, we note that any regularization procedure on the encoder could be thought of as a
modification of the variational family used in the variational approximation.

The same argument is true about architectural choices.
We used a convolutional encoder, as we expect it to be a good match with the deconvolutional generative models under study, but the \gilbo/ would still be valid
if we used an MLP or any other architecture.
The computed \gilbo/ may be more or less tight depending on such choices, though --
the architectural choices for the encoder are a form of inductive bias and should be made in a problem-dependent manner just like any other architectural choice.

\subsection{Finiteness of the Generative Information}
\label{sec:finite}

The generative mutual information is only infinite if the generator network is not only deterministic, but is also invertible.
Deterministic many-to-one functions can have finite mutual informations between their inputs and outputs.
Take for instance the following: $p(z) = \mathcal{U}[-1,1]$, the prior being uniform
from -1 to 1, and a \emph{generator} $x = G(z) = \textrm{sign}(z)$ being the sign function (which is $C^\infty$ almost everywhere),
for which $p(x|z) = \delta(x - \textrm{sign}(z))$ the conditional distribution of $x$ given $z$ is the delta function concentrated on the sign of $z$.

\begin{equation}
	p(x,z) = p(x|z) p(z) = \frac 1 2 \delta(x - \textrm{sign}(z))
	\qquad
	p(z) = \int_{-1}^1 dx\, p(x,z) = \frac 1 2 \delta(z - 1) + \frac 1 2 \delta(z + 1)
\end{equation}

\begin{align}
	I(X;Z) &= \int dx\, dz\, p(x,z) \log \frac{p(x,z)}{p(x)p(z)} \\
 &= \int_{-1}^1 dx\, \int_{-1}^1 dz\, \frac 1 2 \delta(x - \textrm{sign}(z)) \log \frac{ \delta(x - \textrm{sign}(z)) }{ \frac 1 2 \delta(z - 1) + \frac 1 2 \delta(z + 1) } \\
	&= \left[ \frac 1 2 \log 2  \right]_{z=-1} + \left[\frac 1 2 \log 2 \right]_{z=1} = \log 2 = 1 \textrm{bit}
\end{align}
In other words, the deterministic function $x = \textrm{sign}(z)$ induces a mutual
information of 1 bit between $X$ and $Z$.  This makes sense when interpreting the
mutual information as the reduction in the number of yes-no questions needed to specify
the value.
It takes an infinite number of yes-no questions to precisely determine a real
number in the range $[-1, 1]$, but if we observe the sign of the value, it
takes one fewer question (while still being infinite) to determine.

Even if we take $Z$ to be a continuous real-valued random variable on the range
$[-1, 1]$, if we consider a function $x = \texttt{float}(z)$ which casts that
number to a float, for a 32-bit float on the range $[-1, 1]$ the mutual
information that results is $I(X;Z) = 26$ bits (we verified this numerically). In any chain  $Z \rightarrow
\texttt{float}(Z) \rightarrow X$ by the data processing inequality, the mutual
information $I(X;Z)$ is limited by $I(Z;\texttt{float}(Z))=26$ bits (per dimension).
Given that we train neural networks with limited precision arithmetic,
this ensures that there is always some finite mutual information in the representations,
since our random variables are actually discrete, albeit discretized on a very fine grid.

}%

\section{Conclusion}

We've defined a new metric for evaluating generative models, the \gilbo/, and
measured its value on over 3200 models. We've investigated and discussed
strengths and potential limitations of the metric. We've observed that \gilbo/ gives
us different information than is currently available in sample-quality based metrics
like \fid/, both signifying a qualitative difference in the performance of most
\gan/s on \textsc{mnist} versus richer datasets, as well as being able to
distinguish between \gan/s with qualitatively different latent representations
even if they have the same \fid/ score.

On simple datasets, in an information-theoretic sense we cannot distinguish
what \gan/s with the best \fid/s are doing from models that are limited to making
some local deformations of the training set.
On more complicated datasets, \gan/s show a wider array of complexities in their
trained generative models. These complexities cannot be discerned by existing
sample-quality based metrics, but would have important implications for any use of the
trained generative models for auxiliary tasks, such as compression or representation learning.

A truly invertible continuous map from the latent space to the image space
would have a divergent mutual information.  Since \gan/s are implemented as a
feed forward neural network, the fact that we can measure finite and distinct
values for the \gilbo/ for different architectures suggest not only are they
fundamentally not perfectly invertible, but the degree of invertibility is an
interesting signal of the complexity of the learned generative model. Given
that \gan/s are implemented as deterministic feed forward maps, they naturally
want to live at high generative mutual information.

Humans seem to extract only roughly a dozen bits ($\sim 8$ nats) from natural
images into long term memory~\citep{landauer}.  This calls into question the
utility of the usual qualitative visual comparisons of highly complex
generative models.  We might also be interested in trying to train models that
learn much more compressed representations.  \vae/s can naturally target a wide
range of mutual informations~\citep{brokenelbo}. \gan/s are harder to steer.
One approach to make \gan/s steerable is to modify the \gan/ objective and specifically designate a subset of
the full latent space as the informative subspace, as in \citet{infogan}, where
the maximum complexity can be controlled for by limiting the dimensionality of
a discrete categorical latent. The remaining stochasticity in the latent can be
used for novelty in the conditional generations.  Alternatively one could
imagine adding the \gilbo/ as an auxiliary objective to ordinary \gan/
training, though as a lower bound, it may not prove useful for helping to keep
the generative information low. Regardless, we believe it is important to
consider the complexity in information-theoretic terms of the generative models
we train, and the \gilbo/ offers a relatively cheap comparative measure.

We believe using \gilbo/ for further comparisons across architectures,
datasets, and \gan/ and \vae/ variants will illuminate the strengths and
weaknesses of each. The \gilbo/ should be measured and reported when evaluating
any latent variable model. To that end, our implementation is available at
\url{https://github.com/google/compare_gan}.
}%

\subsubsection*{Acknowledgements}
We would like to thank Mario Lucic, Karol Kurach, and Marcin Michalski for the use of their 3200 previously-trained \gan/s and \vae/s and their codebase (described in~\citet{gansequal}),
without which this paper would have had much weaker experiments, as well as for their help adding our \gilbo/ code to their public repository.
We would also like to thank our anonymous reviewers for substantial helpful feedback.

\bibliography{bib}

\begin{thebibliography}{20}
\providecommand{\natexlab}[1]{#1}
\providecommand{\url}[1]{\texttt{#1}}
\expandafter\ifx\csname urlstyle\endcsname\relax
  \providecommand{\doi}[1]{doi: #1}\else
  \providecommand{\doi}{doi: \begingroup \urlstyle{rm}\Url}\fi

\bibitem[Agakov(2006)]{agakov}
Felix~Vsevolodovich Agakov.
\newblock \emph{{Variational Information Maximization in Stochastic
  Environments}}.
\newblock PhD thesis, University of Edinburgh, 2006.

\bibitem[Alemi et~al.(2017)Alemi, Poole, Fischer, Dillon, Saurus, and
  Murphy]{brokenelbo}
Alex Alemi, Ben Poole, Ian Fischer, Josh Dillon, Rif~A. Saurus, and Kevin
  Murphy.
\newblock {Fixing a Broken ELBO}.
\newblock \emph{ICML}, 2017.
\newblock URL \url{https://arxiv.org/abs/1711.00464}.

\bibitem[Arora \& Zhang(2017)Arora and Zhang]{ganbirthday}
Sanjeev Arora and Yi~Zhang.
\newblock {Do GANs actually learn the distribution? An empirical study}.
\newblock \emph{CoRR}, abs/1706.08224, 2017.
\newblock URL \url{http://arxiv.org/abs/1706.08224}.

\bibitem[Belghazi et~al.(2018)Belghazi, Baratin, Rajeshwar, Ozair, Bengio,
  Hjelm, and Courville]{belghazi2018mutual}
Mohamed~Ishmael Belghazi, Aristide Baratin, Sai Rajeshwar, Sherjil Ozair,
  Yoshua Bengio, Devon Hjelm, and Aaron Courville.
\newblock {Mutual Information Neural Estimation}.
\newblock In \emph{International Conference on Machine Learning}, pp.\
  530--539, 2018.

\bibitem[Chen et~al.(2016)Chen, Chen, Duan, Houthooft, Schulman, Sutskever, and
  Abbeel]{infogan}
Xi~Chen, Xi~Chen, Yan Duan, Rein Houthooft, John Schulman, Ilya Sutskever, and
  Pieter Abbeel.
\newblock {InfoGAN: Interpretable Representation Learning by Information
  Maximizing Generative Adversarial Nets}.
\newblock In \emph{NIPS}, 2016.
\newblock URL \url{https://arxiv.org/pdf/1606.03657.pdf}.

\bibitem[{Danihelka} et~al.(2017){Danihelka}, {Lakshminarayanan}, {Uria},
  {Wierstra}, and {Dayan}]{iwc}
I.~{Danihelka}, B.~{Lakshminarayanan}, B.~{Uria}, D.~{Wierstra}, and
  P.~{Dayan}.
\newblock {Comparison of Maximum Likelihood and GAN-based training of Real
  NVPs}.
\newblock \emph{arXiv 1705.05263}, 2017.
\newblock URL \url{https://arxiv.org/abs/1705.05263}.

\bibitem[Gao et~al.(2017)Gao, Kannan, Oh, and Viswanath]{inceptionscore}
Weihao Gao, Sreeram Kannan, Sewoong Oh, and Pramod Viswanath.
\newblock {Estimating Mutual Information for Discrete-Continuous Mixtures}.
\newblock In \emph{Neural Information Processing Systems}, 2017.
\newblock URL
  \url{http://papers.nips.cc/paper/7180-estimating-mutual-information-for-discrete-continuous-mixtures.pdf}.

\bibitem[Goodfellow et~al.(2014)Goodfellow, Pouget-Abadie, Mirza, Xu,
  Warde-Farley, Ozair, Courville, and Bengio]{gan}
Ian Goodfellow, Jean Pouget-Abadie, Mehdi Mirza, Bing Xu, David Warde-Farley,
  Sherjil Ozair, Aaron Courville, and Yoshua Bengio.
\newblock {Generative Adversarial Nets}.
\newblock In \emph{Neural Information Processing Systems}, 2014.
\newblock URL \url{https://arxiv.org/abs/1406.2661}.

\bibitem[{Heusel} et~al.(2017){Heusel}, {Ramsauer}, {Unterthiner}, {Nessler},
  and {Hochreiter}]{fid}
M.~{Heusel}, H.~{Ramsauer}, T.~{Unterthiner}, B.~{Nessler}, and
  S.~{Hochreiter}.
\newblock {GANs Trained by a Two Time-Scale Update Rule Converge to a Local
  Nash Equilibrium}.
\newblock \emph{arXiv 1806.08500}, 2017.
\newblock URL \url{https://arxiv.org/abs/1806.08500}.

\bibitem[Im et~al.(2018)Im, Ma, Taylor, and Branson]{im2018quantitatively}
Daniel~Jiwoong Im, He~Ma, Graham Taylor, and Kristin Branson.
\newblock {Quantitatively evaluating GANs with divergences proposed for
  training}.
\newblock In \emph{International Conference on Learning Representations}, 2018.
\newblock URL \url{https://arxiv.org/abs/1803.01045}.

\bibitem[Kingma \& Ba(2015)Kingma and Ba]{adam}
Diederik Kingma and Jimmy Ba.
\newblock {Adam: A method for stochastic optimization}.
\newblock In \emph{International Conference on Learning Representations}, 2015.
\newblock URL \url{https://arxiv.org/abs/1412.6980}.

\bibitem[Kingma \& Welling(2014)Kingma and Welling]{vae}
Diederik~P Kingma and Max Welling.
\newblock {Auto-encoding variational Bayes}.
\newblock In \emph{International Conference on Learning Representations}, 2014.
\newblock URL \url{https://arxiv.org/abs/1312.6114}.

\bibitem[Landauer(1986)]{landauer}
Thomas~K. Landauer.
\newblock {How much Do People Remember? Some Estimates of the Quantity of
  Learned Information in Long‐term Memory}.
\newblock \emph{Cognitive Science}, 10\penalty0 (4):\penalty0 477--493, 1986.
\newblock \doi{10.1207/s15516709cog1004\_4}.
\newblock URL
  \url{https://onlinelibrary.wiley.com/doi/abs/10.1207/s15516709cog1004_4}.

\bibitem[Lipton \& Tripathi(2017)Lipton and Tripathi]{precise}
Zachary~C Lipton and Subarna Tripathi.
\newblock {Precise Recovery of Latent Vectors From Generative Adversarial
  Networks}, 2017.
\newblock URL \url{https://arxiv.org/abs/1702.04782}.

\bibitem[{Lucic} et~al.(2017){Lucic}, {Kurach}, {Michalski}, {Gelly}, and
  {Bousquet}]{gansequal}
M.~{Lucic}, K.~{Kurach}, M.~{Michalski}, S.~{Gelly}, and O.~{Bousquet}.
\newblock {Are GANs Created Equal? A Large-Scale Study}.
\newblock \emph{arXiv 1711.10337}, 2017.
\newblock URL \url{https://arxiv.org/abs/1711.10337}.

\bibitem[Marsh(2013)]{content}
Charles Marsh.
\newblock {Introduction to Continuous Entropy}, 2013.
\newblock URL
  \url{http://www.crmarsh.com/static/pdf/Charles_Marsh_Continuous_Entropy.pdf}.

\bibitem[Oord et~al.(2018)Oord, Li, and Vinyals]{cpc}
Aaron van~den Oord, Yazhe Li, and Oriol Vinyals.
\newblock Representation learning with contrastive predictive coding.
\newblock \emph{arXiv preprint arXiv:1807.03748}, 2018.

\bibitem[{Talts} et~al.(2018){Talts}, {Betancourt}, {Simpson}, {Vehtari}, and
  {Gelman}]{sbc}
S.~{Talts}, M.~{Betancourt}, D.~{Simpson}, A.~{Vehtari}, and A.~{Gelman}.
\newblock {Validating Bayesian Inference Algorithms with Simulation-Based
  Calibration}.
\newblock \emph{arXiv 1804.06788}, April 2018.
\newblock URL \url{https://arxiv.org/abs/1804.06788}.

\bibitem[{Tishby} \& {Zaslavsky}(2015){Tishby} and {Zaslavsky}]{tishby}
N.~{Tishby} and N.~{Zaslavsky}.
\newblock {Deep Learning and the Information Bottleneck Principle}.
\newblock \emph{arXiv 1503.02406}, 2015.
\newblock URL \url{https://arxiv.org/abs/1503.02406}.

\bibitem[Wu et~al.(2017)Wu, Burda, Salakhutdinov, and
  Grosse]{wu2016quantitative}
Yuhuai Wu, Yuri Burda, Ruslan Salakhutdinov, and Roger Grosse.
\newblock On the quantitative analysis of decoder-based generative models.
\newblock In \emph{International Conference on Learning Representations}, 2017.
\newblock URL \url{https://arxiv.org/abs/1611.04273}.

\end{thebibliography}
\bibliographystyle{iclr2018_workshop}

\end{document}